\pdfoutput=1
\documentclass[11pt]{article}

\usepackage[]{acl}
\usepackage{times}
\usepackage{latexsym}

\usepackage{listings}

\usepackage{multirow}
\usepackage{comment}
\usepackage[caption=false]{subfig}
\usepackage{multicol}
\usepackage{multirow}
\usepackage{booktabs}
\usepackage{xspace}
\usepackage{cleveref}
\usepackage{tabularx}
\usepackage[linewidth=1pt]{mdframed}
\usepackage{graphicx}
\usepackage{subcaption}
\usepackage[]{caption}
\usepackage{float}

\definecolor{c1}{HTML}{4e79a7}%
\definecolor{c2}{HTML}{f28e2b}%
\definecolor{c3}{HTML}{009E73}%
\definecolor{c4}{HTML}{56B4E9}%
\definecolor{c5}{HTML}{CC79A7}%
\definecolor{c6}{HTML}{E69F00}%
\definecolor{c7}{HTML}{844E4D}%
\definecolor{c8}{HTML}{2D512A}%

\definecolor{oorange}{HTML}{d95f02}
\definecolor{bblue}{HTML}{7570b3}
\definecolor{ggreen}{HTML}{1b9e77}
\definecolor{ppurple}{HTML}{e37fbb}
\definecolor{lgreen}{HTML}{9CD24A}
\definecolor{yyellow}{HTML}{FFD52D}
\definecolor{ggold}{HTML}{E1BC89}
\definecolor{ggray}{HTML}{AAAAAA}

\usepackage{siunitx}
\newcommand{\round}[2]{\num[round-mode=places,round-precision=#1]{#2}}

\usepackage[T1]{fontenc}
\usepackage[utf8]{inputenc}

\usepackage{microtype}
\usepackage{twemojis}
\newcommand*\samethanks[1][\value{footnote}]{\footnotemark[#1]}
\title{Your Large Language Models Are Leaving Fingerprints}

\author{
  Hope McGovern \thanks{~ Corresponding Author. \\ Email: \texttt{hope.mcgovern@cl.cam.ac.uk}.} \thanks{~ Work done while at Grammarly.} \\ Cambridge Computer Lab \\ \And
  Rickard Stureborg\samethanks \\ Duke University \\ Grammarly \\ \And
  Yoshi Suhara\samethanks \\ NVIDIA \\ \And
  Dimitris Alikaniotis \\ Grammarly
 }

\begin{document}

\maketitle
\begin{abstract}

It has been shown that finetuned transformers  and other supervised detectors effectively distinguish between human and machine-generated text in some situations \cite{li_deepfake_2023}, 
but we find that even simple classifiers on top of n-gram and part-of-speech features can achieve very robust performance on both in- and out-of-domain data.
To understand how this is possible, we analyze machine-generated output text in five datasets,
finding that LLMs possess unique \textit{fingerprints}
which manifest as slight differences in the frequency of certain lexical and morphosyntactic features.
We show how to visualize such fingerprints, describe how they can be used to detect machine-generated text, and find that they are even robust across textual domains.
We find that fingerprints are often persistent across models in the same model family (e.g. llama-13b vs. llama-65b) and that models fine-tuned for chat are easier to detect than standard language models, indicating that LLM fingerprints may be directly induced by the training data.
% 
% We characterize the fingerprints present in publicly available machine-generated text detection corpora and show that they are persistent across domains and are similar when compared to models within the same model family (e.g. Flan-t5-small and Flan-t5-xxl).
%
% Instruction tuning seems to play an important role in inducing this fingerprint: i.e. models fine-tuned for chat are easier to detect than standard language models.
% %
% We show that the existence of these fingerprints renders LLM-text detection analogous to traditional author identification (AID) and thereby suggest that existing AID methods may be a fruitful avenue for deepfake detection.
% All code is available at \url{www.github.com/anonymous/repo}.

\end{abstract}

\section{Introduction}
% State-of-the-art l
Large language models (LLMs) produce text 
% that is fluent, coherent-sounding, and 
often indistinguishable from human-authored text to human judges \cite{clark_all_2021}. 
% why machine-generated text detection is important
This unfortunately allows potential misuses such as academic plagiarism \cite{westfall_educators_nodate} and the dissemination of disinformation \cite{barnett_chatgpt_nodate},
which has therefore prompted interest in generated text detection (GTD).
% The perceived high-quality output of such language models in conjunction with their public availability render the task of generated text detection (GTD) increasingly important as more recent
%  headlines have highlighted 
% overview of existing, popular methods, and their known weak points
% Many existing methods for machine-generated text detection for the task rely upon token probabilities of a given sequence in order to make a prediction, rather than the text itself. 
% This includes perturbation methods such as [cite DetectGPT], and perplexity-based predictions, such as [GPTZero].
% These approaches, known as white-box  methods, are known to be highly susceptible to adversarial attacks, both advanced [cite deepfake, OUTFOX] and simple [evade detection with a single space]. 
% Making headway on this problem requires, among other things, reliable and representative datasets for benchmarking detection performance. 
We conduct linguistic analysis on five popular published datasets for GTD, showing that the machine-generated content in each shows linguistic markers in aggregate which make it relatively easy to separate it from human content.

These discrepancies, which we call a model's ``fingerprint'', are consistent enough \textit{across domains} and \textit{within model families} that we find we can treat each LLM as if it were a unique author with a distinct writing style.  
To do so, we use two well-founded methods from the field of Author Identification (AID) for a closed set of authors: one using handcrafted n-gram features and another using neural features extracted from pre-trained BERT embeddings, and training a simple machine learning classifier on those features.

\begin{figure*}[t]
    \centering
    \includegraphics[width=\linewidth]{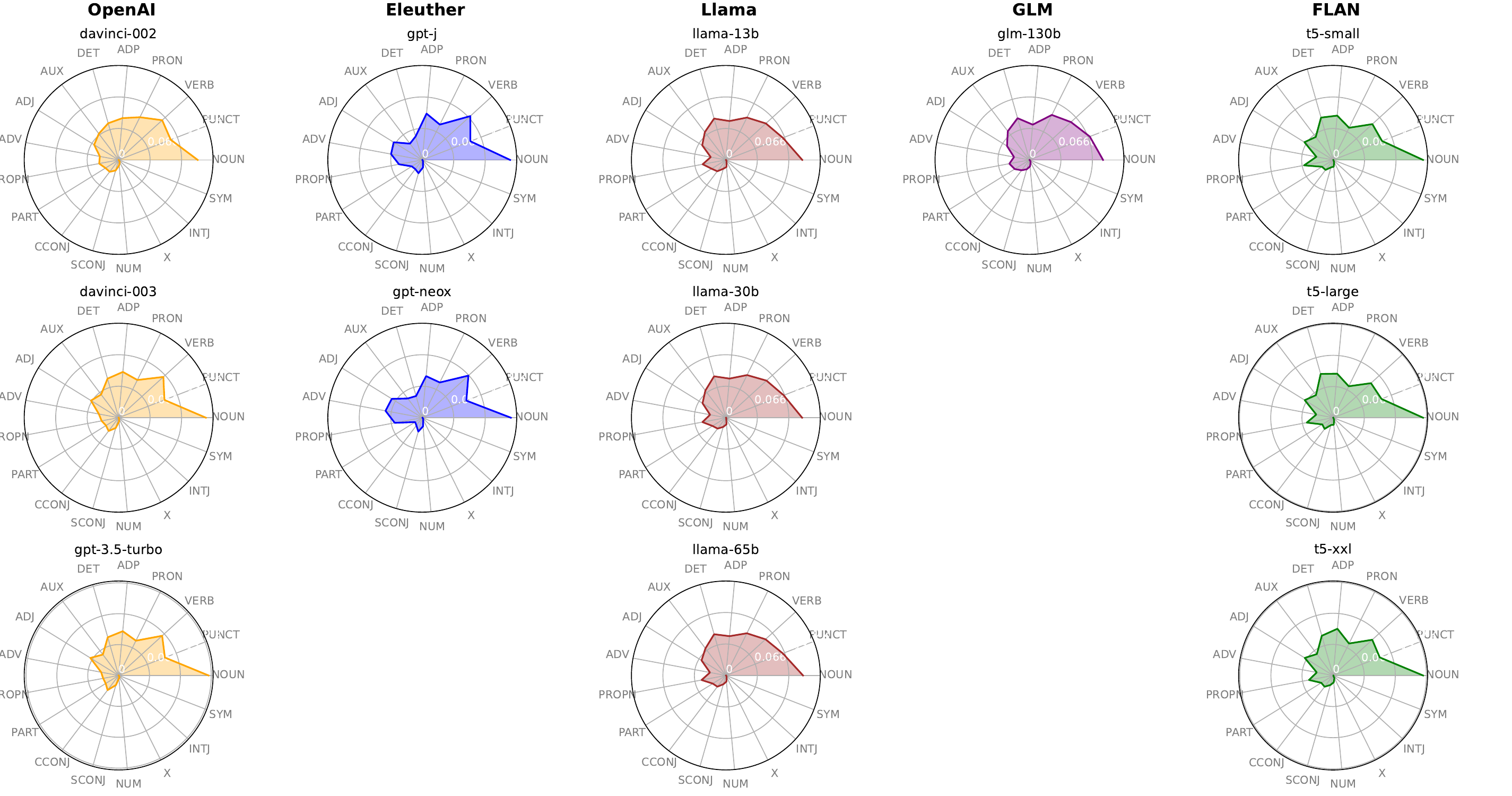}
    \caption{\textbf{Visualization of the fingerprints.} We plot frequencies of each part-of-speech (POS) class from the output of several models, sorted by model family. Within each family, the shapes (distributions) look mostly similar regardless of model size. Each radial plot is shown at the same $0\%$ to $20\%$ frequency scale, with POS tags sorted from most to least common among human-written outputs. Jagged/bumpy shapes indicate the fingerprint is more distinct from human distributions. POS is just one component of the full `fingerprint' we investigate.}
    \label{fig:radial_fingerprints}
\end{figure*}

% \autoref{tab:ngrams_rule} compares the best-reported classifier models from four recent papers which release labeled datasets for GTD: \cite{li_deepfake_2023, hc3, Verma2023GhostbusterDT, outfox} against a GradientBoost \cite{gradientboost} classifier with a combination of character-, word- and POS-n-gram features.\footnote{More implementation details may be found in the Appendix.}

\begin{table}[h]
\footnotesize
    \centering
    \begin{tabular}{lllll} \toprule
        & \multicolumn{2}{c}{Best Reported Model} & \multicolumn{2}{c}{GradientBoost}\\ \cmidrule(lr){2-3} \cmidrule(lr){4-5}
         Paper &  F1&  AUROC & F1 & AUROC \\ \midrule
         Deepfake&  --&   0.99& 94.7& 94.3\\
         HC3&  99.82&   --& \round{1}{96.70583298363408}& \round{1}{99.61814104298692}\\
         Ghostbuster&  99.91&   1.00 & 98 & 98\\ 
         OUTFOX& 96.9& -- & 98.7 & 98.7\\ \bottomrule
    \end{tabular}
    \caption{\textbf{Best reported classifier performances (Deep neural networks) versus a decision-tree model with n-gram features.}
    Best-reported classifier models are from four recent papers which release labeled datasets for GTD: \cite{li_deepfake_2023, hc3, Verma2023GhostbusterDT, outfox}. The GradientBoost \cite{gradientboost} classifier uses a combination of character-, word- and POS-n-gram features.
    GradientBoost models are able to achieve impressively comparable accuracy, even outperforming the best reported model on the OUTFOX benchmark.
    % \rich{is this true??? what was the best model on outfox} \hope{yes, their best model was their OUTFOX detector. Supervised classification on ChatGPT generated essays was 96.9 F1}
    }
    \label{tab:ngrams_rule}
\end{table}

As shown in \autoref{tab:ngrams_rule}, the performance of the simple classifier is surprisingly comparable to more complex neural methods, even in a multi-class setting -- successfully distinguishing between, e.g. human-, ChatGPT-, and LLaMA-generated text (cf. \autoref{tab:multiclass}). It also proves robust in cross-domain experiments (cf. \autoref{fig:deepfake_ood}) and to some adversarial attacks which hinder other detectors (\autoref{tab:adversarial}). Furthermore, we present evidence that instruction-tuning and prompting can manipulate this fingerprint, but not remove it (cf. \autoref{fig:chat_vs_base_POS_distrs}).

In this paper, we empirically uncover and characterize the fingerprints of individual and families of LLMs through a series of comprehensive analyses, and present a new perspective of LLM-content detection as authorship identification.

% Our contributions are fourfold: (1) we demonstrate that LLMs exhibit particular writing styles ("fingerprints"), which are captured by simple, text-based features, and propose (2) a novel, RLHF-based method for removing these fingerprints, thereby producing a language model that produces text with a lexical distribution similar to that of an aggregate of human authors.
%we propose an effective method of removing LLM ``fingerprints'', thereby producing a language model that produces text with lexical distribution similar to that of an aggregate of human authors. 
% We then use this model to (3) produce a machine-generated text detection dataset, which proves challenging for existing detection methods, and (4) release a classifier trained on our dataset, which provides a new and robust benchmark for machine-generated text detection.
%\yoshi{The three contributions look good. Do we think adding ``We show that LLMs have particular writing styles so that simple text-based features can be used to detect machine-generated text accurately on the existing academic benchmarks'' or something like that as another contribution? In other words, I feel like *our finding* of ``Existence of LLM fingerprints'' should be explicitly stated.}

% \input{sections/2_fingerprints}
% \input{sections/3_data}
% \input{sections/4_models}
\section{Methodology}
\label{sec:experiments} 

\subsection{Fingerprint Features}
% Well-substantiated in AID literature is the idea that individual human authors belie their identity in their writing by the use of consistent lexical, syntactical, and grammatical choices \cite{Nini_2023}. We seek to determine whether those same surface level features capture meaningful differences in generated content from individual LLMs. In other words, is GTD a multi-class problem (human vs. ChatGPT vs. Cohere), or is it always more useful to think of GTD as a binary problem (human vs. machine)?

We use three feature sets: word n-grams ($n \in [2,4]$), which we expect to be useful in capturing domain-specific vocabulary, but also in capturing function words, which are known to be highly effective for authorship identification;
character n-grams ($n \in [3,5]$), which we intuitively expect to capture subword information broadly aligning with the byte-pair encoding (BPE)  tokenization method of many models; and part-of-speech (POS) n-grams ($n \in [2,4]$), which should capture domain-agnostic information about writing style.
% 
% For selected experiments, we also consider neural features extracted from a pre-trained language model (i.e. BERT). Neural features are less intuitive and explainable, but \citet{fabien-etal-2020-bertaa} showed that deep contextualised embeddings and a machine learning classifier are effective for AID. We follow \citet{fabien-etal-2020-bertaa}'s method of extracting neural features as they are, with no fine-tuning. Our goal is to use the simplest effective feature set, rather than exhaustively fine-tuning the feature space. This differs from \citet{Petukhova2024PetKazAS}, who use a combination of neural and linguistic features, fine-tuning the neural embeddings beforehand. While both works underline the utility of linguistic features for GTD, our work is interested in pinpointing to what extent LLM detection is really author identification. Where otherwise unspecified in a given experiment, linguistic features are the default.
% We determine n-gram lengths and feature combinations by optimizing for the GTD task, and explore the importance of n-gram lengths in an ablation study (\autoref{apx:additional_exp}).
% We determine n-gram lengths and feature combinations by optimizing for the GTD task, and explore the importance of individual feature sets in an ablation study (\autoref{apx:additional_exp}).

\subsection{Classifiers}
We use a GradientBoost classifier implemented in the Sklearn library \cite{scikit-learn}. The hyperparameters for the classifier were found through grid search, though no extensive hyperparameter sweeps were carried out; this classifier works well out-of-the-box\footnote{Further hyperparameter tuning could improve classifier performance, but we are primarily interested in exploring why such a simple classifier performs well in the first place.}.
% Gradient boosting classification is a machine learning technique where decision trees are built sequentially, each tree correcting errors made by the previous one. It combines the strengths of weak learners to form a strong predictive model by minimizing a loss function through gradient descent\footnote{While we use GradientBoost for all reported experiments in this work, 
Initial experiments used a range of ML classifiers, including SVC and logistic regression. These exhibited close or similar performance on our data.
% }.
% For all ML classifier experiments, training examples are capped to an upper limit of 5000 data points to prevent overfitting. 
If the classes are very imbalanced, we downsample the majority class in order to have a balanced dataset ($n = 5000$).

% We report classifier performance with standard metrics such as \textbf{F1} and \textbf{AUROC} (area under the ROC curve), as well as recall scores for both human (\textbf{HumanRec}) and machine (\textbf{MachineRec)} classes, and their average (\textbf{AvgRec}).

\subsection{Data}
We use five publicly available machine-generated text detection datasets for fingerprint analysis as well as training data for supervised sequence classifiers: OUTFOX \cite{outfox}, DeepfakeTextDetect \cite{li_deepfake_2023}, the Human Comparison Corpus \cite{hc3}, Ghostbuster \cite{Verma2023GhostbusterDT}, and the M4 dataset \cite{wang-etal-2024-m4}.
We refer to these as ``Outfox'',``Deepfake'', ``HC3'', ``Ghostbuster'', and ``M4'' in this work, respectively. 
A summary of all datasets used, including their domain coverage and underlying base model(s), may be seen in \autoref{tab:data_splits}. 
\section{Results and Analysis}
\label{sec:results}

% INTRODUCTION TO RESULTS. Don't change.
We conduct a series of analyses of LLM fingerprints, finding they are
    predictive of which model authored a text,
    consistent across domains and within model families (even to adversarial attacks), 
    and susceptible to modification by fine-tuning.
Additional results are presented in the Appendix.
% Full list of results, along with an extensive discussion section, are presented in the Appendix.

% FLUFF DESCRIPTION OF RADIAL PLOTS. FEEL FREE TO CHANGE.
We visualize fingerprints by looking at the difference of distribution in various linguistic properties. In \autoref{fig:radial_fingerprints}, we report part-of-speech tag distributions of data generated by different models on the same deepfake data domains\footnote{We choose to report POS results in the main paper as it directly maps to one feature set for our classification experiments, whereas we do not directly use named entity categories, constituency types, or top-$k$ words as features.}.  
In \autoref{apx:fingerprint} we also include analysis from named entity tags, constituency types, and top-\textit{k} most frequent tokens. The strength of the fingerprint is more evident across some axes than others, and there are, of course, more dimensions of linguistic analysis that could theoretically be applied to uncover model fingerprints.

Distinct patterns emerge when comparing the fingerprint of models within the same family compared to models of different families. The degree of similarity within families can vary between families; for example, LLaMA models exhibit a particularly uniform fingerprint across model sizes, while BigScience models (cf. \autoref{apx:fingerprint}) look markedly different.
% \rich{Feel free to cut some of the above text, if needed.}

\subsection{Author Identification}
We find that fingerprints are useful for not just GTD, but also to predict which model generated a given text.
\autoref{tab:multiclass} shows that in a multiclass classification setting, these n-gram features allow strong performance for author identification (AID).
This aligns well with previous work in AID where linguistic and stylometric features have been proven highly effective time and time again \cite{He2024}.

\begin{table}
\footnotesize
    \centering
    \begin{tabular}{lll} \toprule
    % \multicolumn{3}{c}{\textbf{Multiclass Classification F1 Score}}\\ \midrule
    \textbf{Dataset} & \textbf{Provenance} & \textbf{F1} \\ \midrule
    
    Ghostbuster & Human & \round{3}{0.934010152284264}\\
                &   ChatGPT & \round{3}{0.9595959595959594}  \\
                & Flan T5 & \round{3}{0.9268292682926829}\\
                \midrule
                & \textbf{Average} & 0.940 \\ \midrule
    Outfox & Human & \round{3}{0.8773584905660379}\\
            & ChatGPT & \round{3}{0.9361702127659574}\\
            & Claude & \round{3}{0.92}\\ \midrule
            & \textbf{Average} & 0.911 \\
            \bottomrule
    \end{tabular}
    \caption{\textbf{F1 scores for each class as the positive class after training under a multiclass classification setting.}
    Note that even for top models ChatGPT and Claude, our simple n-gram based classifier performs very well (0.936 and 0.920 on the Outfox data). To compare with binary classification results, F1 scores are computed for each class by setting that class to be the `positive'.}
    % \textbf{Multiclass Classification results:} We show that fingerprints are sufficient not only to separate `human' from `machine', but to perform accurate multi-class classification, distinguishing between texts produced by different models which are fed the same prompt (the first 30 tokens of the ground truth human response). Here we show multi-class classification results on the Ghostbuster dataset, which includes samples written by humans, ChatGPT and Claude, and the Outfox dataset, which includes samples written by humans, ChatGPT, and Flan.
    % \rich{a little unclear how you get the F1 score of a single class here.. for the "Flan T5" row for example, do you just pick that as the 'positive' class and then calculate F1 from that? If so, add this sentence to the end of above caption: ``To compare with binary classification results, F1 scores are computed for each class by setting that class to be the `positive'. ''}
    % }
    \label{tab:multiclass}
\end{table}

\begin{figure}[h]
    \centering
    \includegraphics[width=0.9\linewidth]{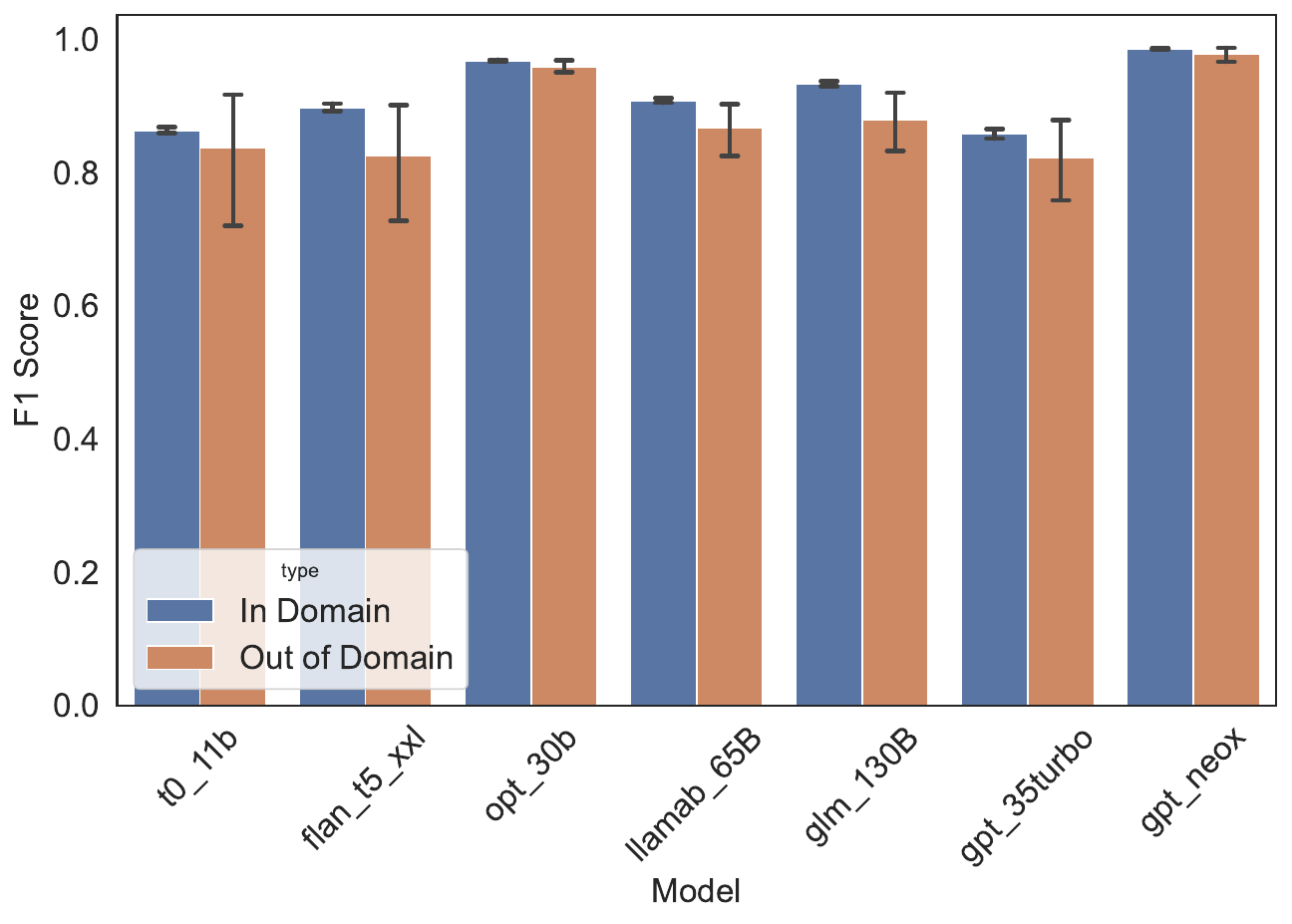}
    \caption{\textbf{F1 score of GTD on in-domain versus out-of-domain test sets for the largest model of each model family in the Deepfake benchmark.} We find no statistically significant drop in performance when testing on these 7 models' outputs. 95\% confidence intervals are computed through bootstrap sampling at $n=10,000$.}
    \label{fig:deepfake_ood}
\end{figure}

\subsection{Robustness}
\begin{figure}[ht]
    \centering
        \includegraphics[width=\linewidth]{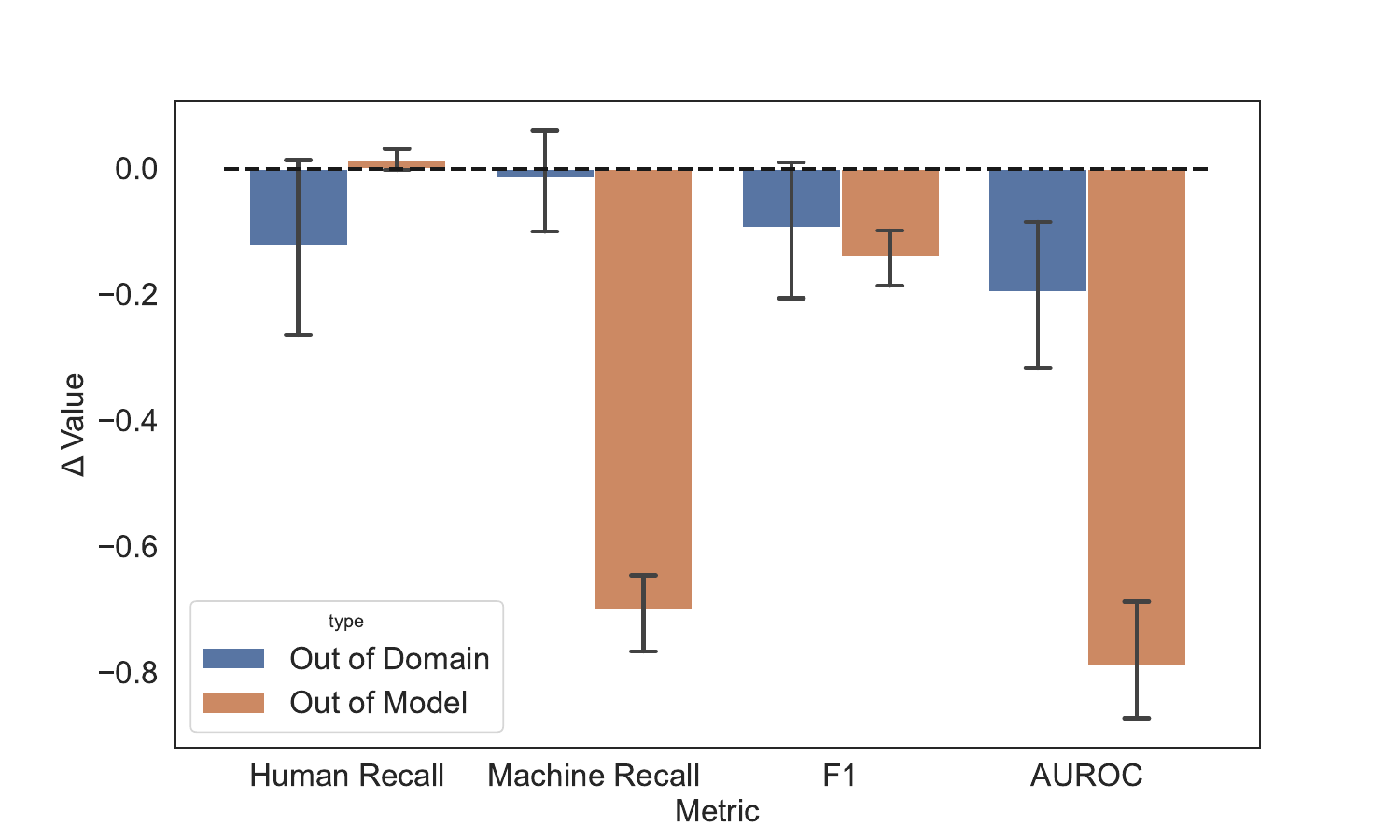}
        \caption{\textbf{Average drop in performance on various metrics when testing on out-of-domain text (blue) versus a held-out generative model (brown).}
        Note that recall of the machine-generated text drops significantly when testing on an unseen model's output, while changing the domain has no impact on this metric.}
        % \textbf{Models exhibit individual writing styles which are more similar across domains than across models.} We report the average drop in performance of a GradientBoost classifier trained on Deepfake data. We sample text from the largest model in each model family: LLaMa 65B, BigScience 11B, Flan T5 XXL, OPT 30B, GLM 130B, GPT-3.5-turbo, and GPT NeoX 20B. In 10 independent trials, we train a classifier on one model and compare its performance on the in-domain test set to: (1) data from the same model but in a held-out domain, and (2) data from a different model but same domains present in the train set. We then use bootstrapping to calculate the 95\% confidence interval, sampling with replacement 10,000 times. Machine recall meaningfully drops when being evaluated on a different model, but not so on a different domain from the same model; much like how a human author, whether writing about recipes or current events, will have some degree of consistency of style, so too do LLMs.}
        %\small\textsuperscript{a}{this is made possible by the fact that Deepfake data is multi-parallel}
        % \caption{\textbf{LLM writing styles vs. domain}. A selected visualization from the results summarized in \autoref{tab:ood_v_oom}. This plot shows a classifier trained on data from GPT-J-6B, evaluated on held-out data domains and data from BLOOM models. Full results may be seen in the Appendix.}
        \label{fig:ood_v_oom}
\end{figure}

Given the nature of the fingerprint features, one might expect that a shift in domain substantially impacts the performance of GTD using these fingerprints.
However, in \autoref{fig:deepfake_ood} we find no evidence that performance deteriorates for out-of-domain test sets, while \autoref{fig:ood_v_oom} shows a clear deterioration due to out-of-model test sets.
That is, fingerprints are robust across domains but are unique for each model.
Even further, fingerprints are somewhat robust to adversarial attacks \autoref{tab:adversarial}.

\subsection{Altering fingerprints}
\begin{figure}[h]
    \centering
    \includegraphics[width=.95\linewidth]{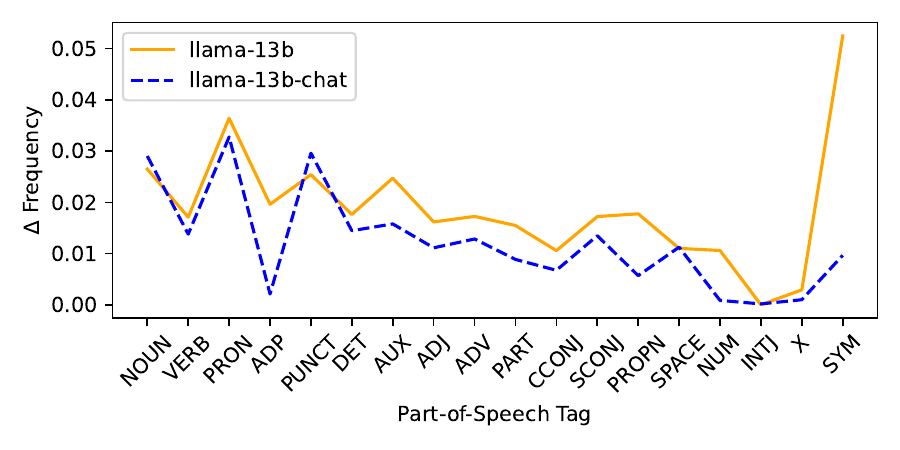}
    \caption{\textbf{Absolute difference in POS tag frequencies as compared with human text}. Chat models are slightly more similar to the frequency profile of humans, but are easier to detect than base models. This demonstrates that fingerprints ``closer'' to human distributions in POS tags does \textit{not} indicate it is less detectable.
    Further, fine-tuning models for chat clearly alters their fingerprint despite no change in model architecture.}
    \label{fig:chat_vs_base_POS_distrs}
\end{figure}

Some adversarial attacks seem to potentially alter fingerprints \autoref{tab:adversarial}, though qualitatively they also change the readability of the final text.
Alternatively, instruction tuning through reinforcement learning or supervised fine-tuning is a potential method to purposefully alter model fingerprints.
\autoref{fig:chat_vs_base_POS_distrs} shows the potential of altering fingerprints by comparing a chat model's fingerprint to the base model.
% \rich{please make note: we need to add F1 scores for chat versus standard models and write them directly into this text. Let's just do that for the rebuttal with reviewers.}

\begin{table}[h]
\footnotesize
\centering
\begin{tabular}{cccc}
\toprule
\textbf{CLS}                                      & \multicolumn{1}{c}{\textbf{Unattacked}} & \multicolumn{1}{c}{\textbf{DIPPER}} & \multicolumn{1}{c}{\textbf{OUTFOX}} \\ \hline
\multicolumn{1}{l|}{Ling + GB}           & \round{3}{0.992}                                             & \round{3}{0.7613200306983884}                       & \round{3}{0.9668615984405458}                       \\
\multicolumn{1}{l|}{Bert + GB}           & \round{3}{0.9919354838709677}                                & \textbf{\round{3}{0.8482758620689654}}                       & \round{3}{0.979757085020243}                       \\
\multicolumn{1}{l|}{Bert + LR}             & \textbf{\round{3}{0.997979797979798}}                                 & \round{3}{0.8111658456486043}                       & \textbf{\round{3}{0.9918032786885246}}                       \\ \hline
\multicolumn{1}{l|}{Deepfake} & \round{3}{0.5493333333333333}                                & \round{3}{0.6609625668449198}                       & \round{3}{0.7639060568603214}                     \\
% \multicolumn{1}{c|}{DetectGPT}           &                                                   &                                          &                                          \\
\multicolumn{1}{l|}{GPTZero}             &  \round{3}{0.4065573770491803}                                                 &  \round{3}{0.6372093023255814}                                        &   \round{3}{0.6919191919191918}    \\  
\bottomrule
\end{tabular}
    \caption{\textbf{F1 scores before and after adversarial attacks by DIPPER and OUTFOX.}
    GB stands for GradientBoost, LR stands for Logistic Regression. 
    % \rich{LR stands for what??}
    Ling are linguistic features, and Bert are BERT-based embeddings as features.
    Note that our models are robust to OUTFOX attacks, while they deteriorate slightly by DIPPER (though still showing relatively strong F1 scores).
    BERT-based features are more robust to DIPPER than our linguistic features.
    % \rich{please read this caption and make sure I've understood this table correctly.}
    % \rich{we need to be ready to defend why the performance increased for deepfake/gptzero after adversarial attacks.... Reviewers will ask. (which is fine, sort of nice to be able to prepare in advance).}
    }
\label{tab:adversarial}
\end{table}

\section{Related Work}

A common approach to machine-generated text detection is to train a supervised binary classifier on labeled data \cite{hc3, outfox, li_deepfake_2023}.
% [OpenAI] released a detector that was a GPT model fine-tuned in this manner; however, it required an input text length of at least 1,000 characters and was eventually retired due to poor performance [cite that news release]. 
% \citet{li_deepfake_2023} proposed a variety of classification testbeds, methodically mixing data generated by 27 different models in a number of domains to rigorously test the ability of supervised classifiers to generalise to unseen domains and unseen models, finding that pre-trained language models perform the best, but still suffer a significant performance drop on out-of-distribution data. 
\citet{li_deepfake_2023} proposed a variety of classification testbeds, finding that pre-trained language models perform the best.
While $n$-gram frequencies have often been used for author identification, only a few recent works examine hand-crafted features or stylometrics in machine-generated text detection \cite{zaitsu_distinguishing_2023}. 
One example is \citet{Gehrmann2019GLTRSD}: a unique system that uses the top-$k$ words to highlight text spans
% likely to be generated by an LLM. 
% GLTR is not a detector itself, per se, as its aim is 
to visually aid humans in the task of spotting AI-written text themselves.

% \textbf{Linguistic Features for LLM-generated text detection}
Some of our findings support recent work: \citet{zaitsu_distinguishing_2023} find stylometric methods effective for detection in Japanese text, even compared with SOTA models. 
% Additionally, \citet{li_deepfake_2023} find that only one of their tested methods is reasonably effective at detecting text generated from unseen models.
Additionally, \citet{Gehrmann2019GLTRSD} have proposed a feature extractor for feature-based, statistical classification for machine-generated text detection, and
\citet{Petukhova2024PetKazAS} finds a combination of fine-tuned neural features and hand-crafted linguistic features effective for GTD on the M4 dataset as part of the SemEval2024 task on machine-generated text detection \cite{semeval2024task8}.

% \textbf{Linguistic Analysis.} To our knowledge, we are the first to develop a theory of unique LLM fingerprints and show that they are more similar within model families than across them. Other work, 
For linguistic analysis, \citet{li_deepfake_2023} analyze their corpus DeepfakeTextDetect, but report differences across POS-tag distributions between human and machine data when considering all models and domains in aggregate as insignificant; however, they do find these distributions begin to diverge when considering a subset of models or domains. We demonstrate that these differences extend to every publicly available machine text detection dataset, prove largely consistent within model families, and are very powerful features for training a robust machine-generated text detection classifier. 
%
% \citet{hc3} also notice and comment upon the differences in POS-tag distribution between their collected ChatGPT and human data, positing that this discrepancy may be the result of the RLHF process.
%
% Unlike these methods, we use an n-gram feature set, rather than stylometric features or surprisal metrics, and a simple machine learning classifier (GradientBoost).
%
% [Ghostbuster] proposes using a series of weaker language models, including $n$-gram models to determine features on which to train an ensemble classifier. 
% The downside of this approach, in particular, is that it requires access to multiple language models for inference, which may be resource- or cost-prohibitive.
%
%
%why we didn't do adversarial methods -- we are less interested in fooling a detector than we are creating data that is truly indistinguishable from that of humans
% \input{sections/8_background}
\section{Conclusion}
% Overall message
We demonstrate that in five popular datasets for machine-generated text detection, n-gram features are highly effective for GTD.
% in-domain, out-of-domain, and multi-class detection 
% for GTD. We show that these feature sets are effective, without any fine-tuning or particular hyperparameter search, because 
We uncover that LLMs have unique writing styles that can be captured in lexical and syntactic features, which we characterize as ``fingerprints'', 
% , percentage discrepancies of human vs LLM POS-tag, named entity tag, constituency tags, and top-k token frequency, 
and show that they are generally unique to a model family.
% , with slight differences across individual models within a family. 
We also show some evidence fingerprints can be modified with further fine-tuning, suggesting they may be removable.
% that authorship identification may be a worthwhile direction for future research into machine-generated text detection.
% We propose an experimental setup based on RLHF that fails to satisfactorily remove evidence of fingerprints, suggesting that fingerprints are the product of a complex interaction between model architecture and training data which may not be easily removed.

\section*{Limitations}
\begin{itemize}
    
    \item \textbf{Text length:} we examine outputs of approximately 300-500 words in length. Shorter texts may be difficult to fingerprint or may not provide enough signal.
    %\item Our methods assume no learned adversarial attacks, which are known to be effective. However, our framework should be more robust than existing ones, since it can be viewed as a form of evading detection. 
    % \item \textbf{Generalization.} Although we have demonstrated general trends in five publicly available datasets, we cannot be certain that our 
    \item \textbf{Prompting.} We do not explore prompting methods in any exhaustive or fine-grained manner, although we do note that we conduct analysis on datasets that have used a variety of prompting methods themselves to collect data. Still, we acknowledge that choice of prompts has been shown to have a significant impact on the output of an LLM.
    \item \textbf{Model choice limitations}: We constrain ourselves to the data and models released as part of text detection corpora, which means that there some very good models we simply did not have the data to test, e.g. GPT4 data.
    \item \textbf{Generation uncertainty}: For our instruction-tuning fingerprint experiment, we only fine-tune 7b and 13b variants of Llama-2. These are relatively small models and there is no guarantee that our methods would work for larger models or different instruction tuning regimes.
    \item \textbf{Reflection on real-world use-case.} Analyzing fingerprints in research benchmark datasets is most likely \textit{not} reflective of the true difficulty of deepfake text detection in the wild. For one thing, people don't tend to use LLMs for writing entire articles/essays, etc. A more likely scenario for, e.g. academic plagiarism, is starting from an LLM generated paragraph and making sentence-level rewrites. As this is analagous to a paraphrase attack like DIPPER \cite{krishna2023paraphrasing}, we expect that it would degrade our classifiers' performance.
\end{itemize}

\section*{Ethics Statement}
This research indicates that detecting machine-generated text is easy.
However, we want to stress that this does \textit{not} necessarily mean machine-detection is a high-confidence task.
Using a single model prediction about one single written text to determine whether or not it was human-written should be evaluated on a different basis than average accuracy, given the potential harms of false positives or false negatives.
For example, teachers may wish to use tools to determine if students have cheated on exams or homework using LLMs.
We discourage teachers from trusting predictions by any classifier until more investigation is done into the confidence models have for any individual text.

% We are aware that this work could contribute
% Our intention with this work is not to make a 

% \newpage
\bibliography{llm_detection}

\begin{thebibliography}{25}
\expandafter\ifx\csname natexlab\endcsname\relax\def\natexlab#1{#1}\fi

\bibitem[{Barnett(2023)}]{barnett_chatgpt_nodate}
Sofia Barnett. 2023.
\newblock \href {https://www.wired.com/story/chatgpt-college-university-plagiarism/} {{ChatGPT} {Is} {Making} {Universities} {Rethink} {Plagiarism}}.
\newblock \emph{Wired}.
\newblock Section: tags.

\bibitem[{Beltagy et~al.(2020)Beltagy, Peters, and Cohan}]{longformer}
Iz~Beltagy, Matthew~E. Peters, and Arman Cohan. 2020.
\newblock \href {https://api.semanticscholar.org/CorpusID:215737171} {Longformer: The long-document transformer}.
\newblock \emph{ArXiv}, abs/2004.05150.

\bibitem[{Cai and Cui(2023)}]{Cai2023EvadeCD}
Shuyang Cai and Wanyun Cui. 2023.
\newblock \href {https://api.semanticscholar.org/CorpusID:259360677} {Evade chatgpt detectors via a single space}.
\newblock \emph{ArXiv}, abs/2307.02599.

\bibitem[{Clark et~al.(2021)Clark, August, Serrano, Haduong, Gururangan, and Smith}]{clark_all_2021}
Elizabeth Clark, Tal August, Sofia Serrano, Nikita Haduong, Suchin Gururangan, and Noah~A. Smith. 2021.
\newblock \href {https://doi.org/10.18653/v1/2021.acl-long.565} {All {That}’s ‘{Human}’ {Is} {Not} {Gold}: {Evaluating} {Human} {Evaluation} of {Generated} {Text}}.
\newblock In \emph{Proceedings of the 59th {Annual} {Meeting} of the {Association} for {Computational} {Linguistics} and the 11th {International} {Joint} {Conference} on {Natural} {Language} {Processing} ({Volume} 1: {Long} {Papers})}, pages 7282--7296, Online. Association for Computational Linguistics.

\bibitem[{Franklin et~al.(2022)Franklin, Maggie, Benner, Rambis, Baffour, Holbrook, Crossley, and ulrichboser}]{feedback-prize-effectiveness}
Alex Franklin, Maggie, Meg Benner, Natalie Rambis, Perpetual Baffour, Ryan Holbrook, Scott Crossley, and ulrichboser. 2022.
\newblock \href {https://kaggle.com/competitions/feedback-prize-effectiveness} {Feedback prize - predicting effective arguments}.

\bibitem[{Friedman(2001)}]{gradientboost}
Jerome~H Friedman. 2001.
\newblock Greedy function approximation: a gradient boosting machine.
\newblock \emph{Annals of statistics}, pages 1189--1232.

\bibitem[{Gehrmann et~al.(2019)Gehrmann, Strobelt, and Rush}]{Gehrmann2019GLTRSD}
Sebastian Gehrmann, Hendrik Strobelt, and Alexander~M. Rush. 2019.
\newblock \href {https://api.semanticscholar.org/CorpusID:182952848} {Gltr: Statistical detection and visualization of generated text}.
\newblock In \emph{Annual Meeting of the Association for Computational Linguistics}.

\bibitem[{Guo et~al.(2023)Guo, Zhang, Wang, Jiang, Nie, Ding, Yue, and Wu}]{hc3}
Biyang Guo, Xin Zhang, Ziyuan Wang, Minqi Jiang, Jinran Nie, Yuxuan Ding, Jianwei Yue, and Yupeng Wu. 2023.
\newblock \href {https://doi.org/10.48550/ARXIV.2301.07597} {How {Close} is {ChatGPT} to {Human} {Experts}? {Comparison} {Corpus}, {Evaluation}, and {Detection}}.
\newblock Publisher: arXiv Version Number: 1.

\bibitem[{He et~al.(2024)He, Lashkari, Vombatkere, and Sharma}]{He2024}
Xie He, Arash~Habibi Lashkari, Nikhill Vombatkere, and Dilli~Prasad Sharma. 2024.
\newblock \href {https://doi.org/10.3390/info15030131} {Authorship attribution methods, challenges, and future research directions: A comprehensive survey}.
\newblock \emph{Information}, 15(3):131.

\bibitem[{Kirchenbauer et~al.(2023)Kirchenbauer, Geiping, Wen, Katz, Miers, and Goldstein}]{watermark}
John Kirchenbauer, Jonas Geiping, Yuxin Wen, Jonathan Katz, Ian Miers, and Tom Goldstein. 2023.
\newblock \href {https://api.semanticscholar.org/CorpusID:256194179} {A watermark for large language models}.
\newblock In \emph{International Conference on Machine Learning}.

\bibitem[{Koike et~al.(2023)Koike, Kaneko, and Okazaki}]{outfox}
Ryuto Koike, Masahiro Kaneko, and Naoaki Okazaki. 2023.
\newblock \href {https://api.semanticscholar.org/CorpusID:260091573} {Outfox: Llm-generated essay detection through in-context learning with adversarially generated examples}.
\newblock \emph{ArXiv}, abs/2307.11729.

\bibitem[{Krishna et~al.(2023)Krishna, Song, Karpinska, Wieting, and Iyyer}]{krishna2023paraphrasing}
Kalpesh Krishna, Yixiao Song, Marzena Karpinska, John Wieting, and Mohit Iyyer. 2023.
\newblock \href {http://arxiv.org/abs/2303.13408} {Paraphrasing evades detectors of ai-generated text, but retrieval is an effective defense}.

\bibitem[{Li et~al.(2023)Li, Li, Cui, Bi, Wang, Yang, Shi, and Zhang}]{li_deepfake_2023}
Yafu Li, Qintong Li, Leyang Cui, Wei Bi, Longyue Wang, Linyi Yang, Shuming Shi, and Yue Zhang. 2023.
\newblock \href {https://doi.org/10.48550/ARXIV.2305.13242} {Deepfake {Text} {Detection} in the {Wild}}.
\newblock Publisher: arXiv Version Number: 1.

\bibitem[{Mitchell et~al.(2023)Mitchell, Lee, Khazatsky, Manning, and Finn}]{mitchell_detectgpt_2023}
Eric Mitchell, Yoonho Lee, Alexander Khazatsky, Christopher~D. Manning, and Chelsea Finn. 2023.
\newblock \href {https://doi.org/10.48550/ARXIV.2301.11305} {{DetectGPT}: {Zero}-{Shot} {Machine}-{Generated} {Text} {Detection} using {Probability} {Curvature}}.
\newblock arXiv.
\newblock Version Number: 2.

\bibitem[{Ouyang et~al.(2022)Ouyang, Wu, Jiang, Almeida, Wainwright, Mishkin, Zhang, Agarwal, Slama, Ray, Schulman, Hilton, Kelton, Miller, Simens, Askell, Welinder, Christiano, Leike, and Lowe}]{rlhf}
Long Ouyang, Jeff Wu, Xu~Jiang, Diogo Almeida, Carroll~L. Wainwright, Pamela Mishkin, Chong Zhang, Sandhini Agarwal, Katarina Slama, Alex Ray, John Schulman, Jacob Hilton, Fraser Kelton, Luke~E. Miller, Maddie Simens, Amanda Askell, Peter Welinder, Paul~Francis Christiano, Jan Leike, and Ryan~J. Lowe. 2022.
\newblock \href {https://api.semanticscholar.org/CorpusID:246426909} {Training language models to follow instructions with human feedback}.
\newblock \emph{ArXiv}, abs/2203.02155.

\bibitem[{Pedregosa et~al.(2011)Pedregosa, Varoquaux, Gramfort, Michel, Thirion, Grisel, Blondel, Prettenhofer, Weiss, Dubourg, Vanderplas, Passos, Cournapeau, Brucher, Perrot, and Duchesnay}]{scikit-learn}
F.~Pedregosa, G.~Varoquaux, A.~Gramfort, V.~Michel, B.~Thirion, O.~Grisel, M.~Blondel, P.~Prettenhofer, R.~Weiss, V.~Dubourg, J.~Vanderplas, A.~Passos, D.~Cournapeau, M.~Brucher, M.~Perrot, and E.~Duchesnay. 2011.
\newblock Scikit-learn: Machine learning in {P}ython.
\newblock \emph{Journal of Machine Learning Research}, 12:2825--2830.

\bibitem[{Petukhova et~al.(2024)Petukhova, Kazakov, and Kochmar}]{Petukhova2024PetKazAS}
Kseniia Petukhova, Roman Kazakov, and Ekaterina Kochmar. 2024.
\newblock \href {https://api.semanticscholar.org/CorpusID:269004629} {Petkaz at semeval-2024 task 8: Can linguistics capture the specifics of llm-generated text?}

\bibitem[{Rafailov et~al.(2023)Rafailov, Sharma, Mitchell, Ermon, Manning, and Finn}]{rafailov_direct_2023}
Rafael Rafailov, Archit Sharma, Eric Mitchell, Stefano Ermon, Christopher~D. Manning, and Chelsea Finn. 2023.
\newblock \href {https://doi.org/10.48550/ARXIV.2305.18290} {Direct {Preference} {Optimization}: {Your} {Language} {Model} is {Secretly} a {Reward} {Model}}.
\newblock Publisher: arXiv Version Number: 2.

\bibitem[{Salinas and Morstatter(2024)}]{Salinas2024TheBE}
Abel Salinas and Fred Morstatter. 2024.
\newblock \href {https://api.semanticscholar.org/CorpusID:266844185} {The butterfly effect of altering prompts: How small changes and jailbreaks affect large language model performance}.
\newblock \emph{ArXiv}, abs/2401.03729.

\bibitem[{Soto et~al.(2021)Soto, Miano, Ordo{\~n}ez, Chen, Khan, Bishop, and Andrews}]{Soto2021LearningUA}
Rafael A.~Rivera Soto, Olivia~Elizabeth Miano, Juanita Ordo{\~n}ez, Barry~Y. Chen, Aleem Khan, Marcus Bishop, and Nicholas Andrews. 2021.
\newblock \href {https://api.semanticscholar.org/CorpusID:243865314} {Learning universal authorship representations}.
\newblock In \emph{Conference on Empirical Methods in Natural Language Processing}.

\bibitem[{Verma et~al.(2023)Verma, Fleisig, Tomlin, and Klein}]{Verma2023GhostbusterDT}
Vivek~Kumar Verma, Eve Fleisig, Nicholas Tomlin, and Dan Klein. 2023.
\newblock \href {https://api.semanticscholar.org/CorpusID:258865787} {Ghostbuster: Detecting text ghostwritten by large language models}.
\newblock \emph{ArXiv}, abs/2305.15047.

\bibitem[{Wang et~al.(2024{\natexlab{a}})Wang, Mansurov, Ivanov, Su, Shelmanov, Tsvigun, Whitehouse, Afzal, Mahmoud, Puccetti, Arnold, Aji, Habash, Gurevych, and Nakov}]{semeval2024task8}
Yuxia Wang, Jonibek Mansurov, Petar Ivanov, Jinyan Su, Artem Shelmanov, Akim Tsvigun, Chenxi Whitehouse, Osama~Mohammed Afzal, Tarek Mahmoud, Giovanni Puccetti, Thomas Arnold, Alham~Fikri Aji, Nizar Habash, Iryna Gurevych, and Preslav Nakov. 2024{\natexlab{a}}.
\newblock {SemEval}-2024 task 8: Multigenerator, multidomain, and multilingual black-box machine-generated text detection.
\newblock In \emph{Proceedings of the 18th International Workshop on Semantic Evaluation}, SemEval 2024, Mexico City, Mexico.

\bibitem[{Wang et~al.(2024{\natexlab{b}})Wang, Mansurov, Ivanov, Su, Shelmanov, Tsvigun, Whitehouse, Mohammed~Afzal, Mahmoud, Sasaki, Arnold, Aji, Habash, Gurevych, and Nakov}]{wang-etal-2024-m4}
Yuxia Wang, Jonibek Mansurov, Petar Ivanov, Jinyan Su, Artem Shelmanov, Akim Tsvigun, Chenxi Whitehouse, Osama Mohammed~Afzal, Tarek Mahmoud, Toru Sasaki, Thomas Arnold, Alham Aji, Nizar Habash, Iryna Gurevych, and Preslav Nakov. 2024{\natexlab{b}}.
\newblock \href {https://aclanthology.org/2024.eacl-long.83} {M4: Multi-generator, multi-domain, and multi-lingual black-box machine-generated text detection}.
\newblock In \emph{Proceedings of the 18th Conference of the European Chapter of the Association for Computational Linguistics (Volume 1: Long Papers)}, pages 1369--1407, St. Julian{'}s, Malta. Association for Computational Linguistics.

\bibitem[{Westfall(2023)}]{westfall_educators_nodate}
Chris Westfall. 2023.
\newblock \href {https://www.forbes.com/sites/chriswestfall/2023/01/28/educators-battle-plagiarism-as-89-of-students-admit-to-using-open-ais-chatgpt-for-homework/} {Educators {Battle} {Plagiarism} {As} 89\% {Of} {Students} {Admit} {To} {Using} {OpenAI}’s {ChatGPT} {For} {Homework}}.
\newblock Section: Careers.

\bibitem[{Zaitsu and Jin(2023)}]{zaitsu_distinguishing_2023}
Wataru Zaitsu and Mingzhe Jin. 2023.
\newblock \href {https://doi.org/10.1371/journal.pone.0288453} {Distinguishing {ChatGPT}(-3.5, -4)-generated and human-written papers through {Japanese} stylometric analysis}.
\newblock \emph{PLOS ONE}, 18(8):e0288453.

\end{thebibliography}
\bibliographystyle{acl_natbib}

\newpage
\appendix
\onecolumn
\section{Discussion}
\label{sec:discussion}
\textbf{A note about prompting.} Different prompting methods have quite a large effect on the output generation and quality of LLMs \cite{Salinas2024TheBE}. While it is possible that different prompting methods could then have a large effect on the measured fingerprint of a single output, at the dataset level, a little bit of prompting variation seems not to be too detrimental. Specifically, every dataset we tested used a combination of prompting techniques -- continuation, topic based, chain-of-thought, etc. and we do still find LLM style markers which are useful for classification. Furthermore, we show that both linguistic and neural features + GB classifier is robust to an adversarial technique that uses different prompts to try to fool a detector, OUTFOX, in \autoref{tab:adversarial}.

\subsection{What to Do About Fingerprints}
The knowledge that LLM fingerprints may be exploited for quick, accurate, explainable classifications we hope will encourage more research into other straightforward and trustworthy methods of machine-text detection. In\autoref{apx:rlhf}, we detail the negative results from a Reinforcement Learning with Human Feedback (RLHF) setup in an attempt to remove the fingerprints. Our inconclusive RLHF experiments raises the question of the intrinsic-ness of the fingerprints and if they can be removed at all. If they cannot, then fingerprints may remain a strong classification signal for the task until other training procedures prove to produce fingerprint-less models. 

Perhaps the most interesting insight from our work is that LLMs may be considered analogous to human authors for the purpose of deepfake detection. We hope that this insight will bridge two research avenues and help to develop a more robust theory of LLM generation.

% and for people
Finally, we encourage those releasing datasets for machine-generated text detection to benchmark against simple, feature-based classifiers.

% \textbf{Future Work}
% More models, more data domains, creating a heuristic for knowing when a detector will work on a certain domain. 

\section{Fingerprint Characterization}
\label{apx:fingerprint}
\begin{figure}[H]
    \centering
    \includegraphics[width=0.6\linewidth]{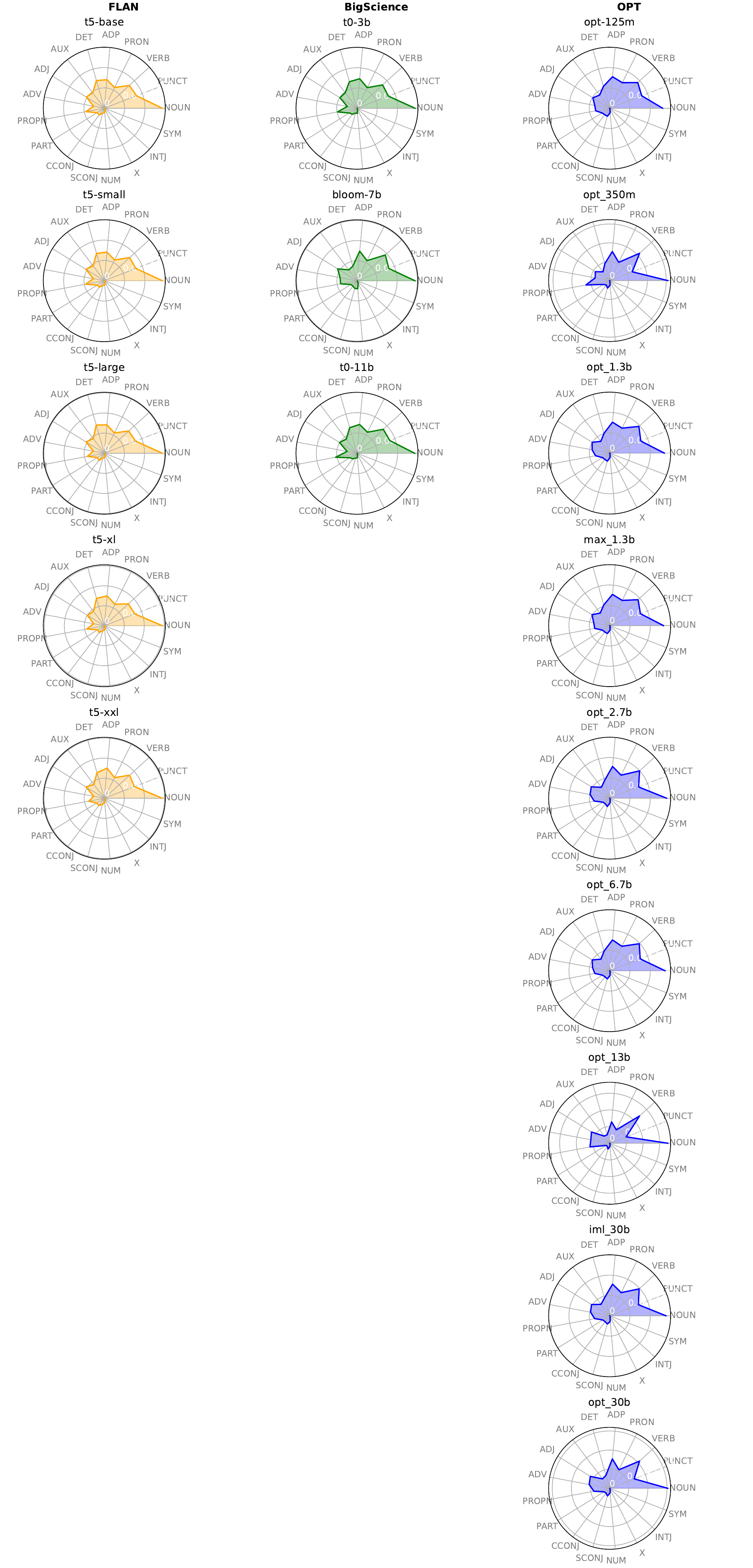}
    \caption{\textbf{Additional visualizations of fingerprints.} Note that the POS tag distributions of OPT models are less similar than we observe within other model families. Further investigations could examine what causes these differences, since model size seems to not play a factor in FLAN models.}
    \label{fig:rest_of_radials}
\end{figure}

\begin{figure}
\begin{tabular}{ccc}
\subfloat[BigScience]{\includegraphics[width = 2 in]{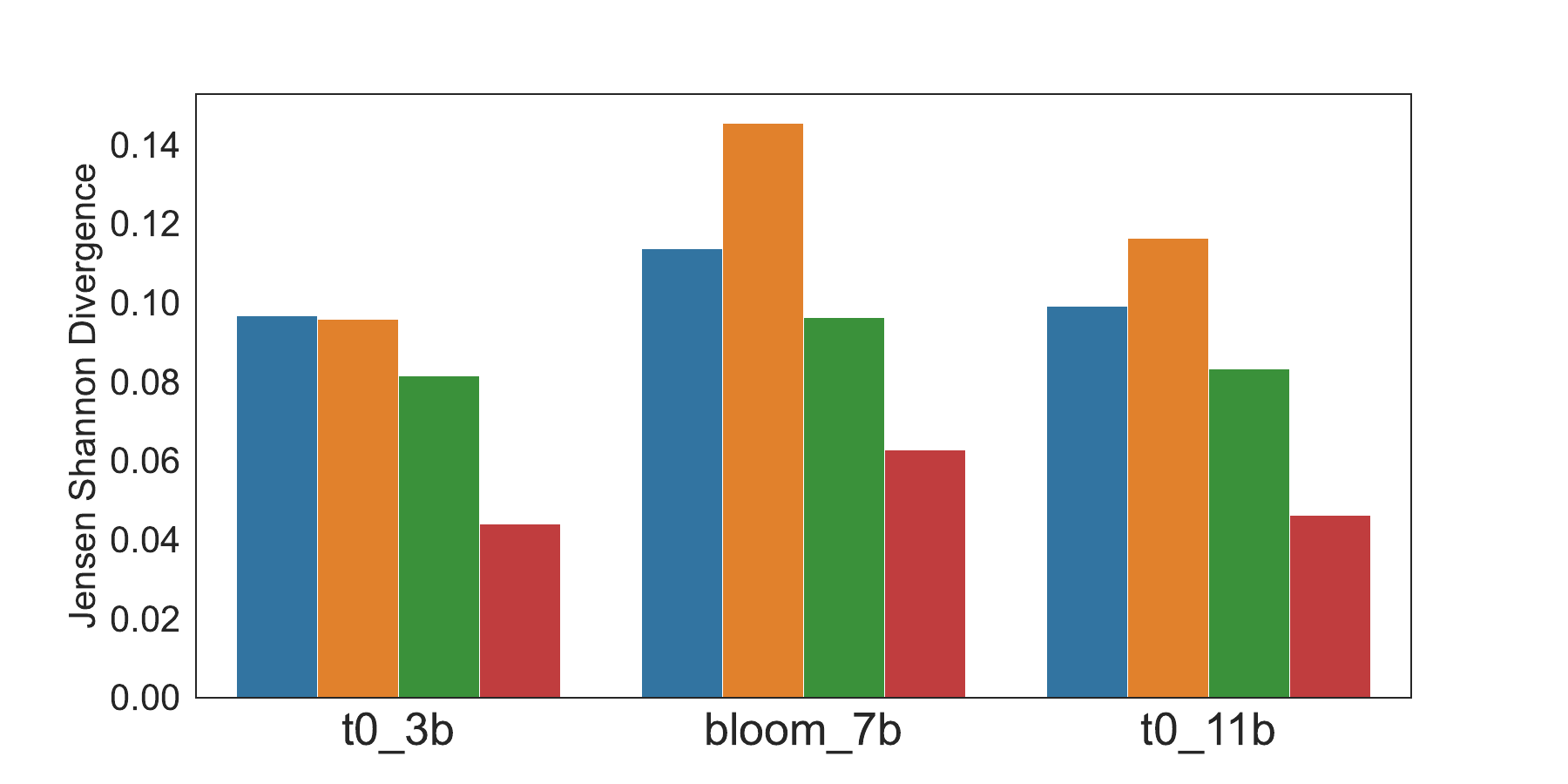}} &
\subfloat[EleutherAI]{\includegraphics[width = 2 in]{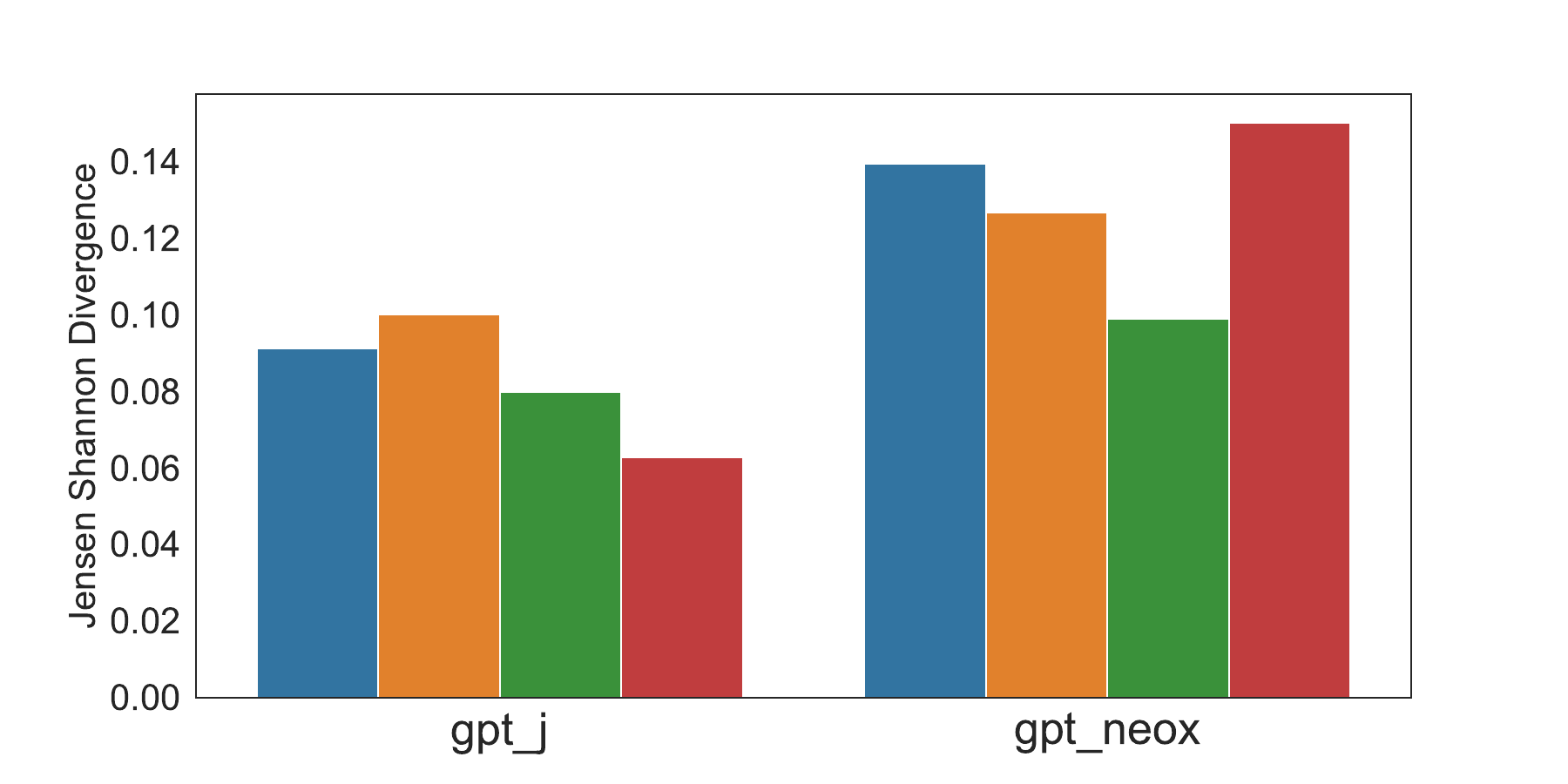}} &
\subfloat[Flan]{\includegraphics[width = 2 in]{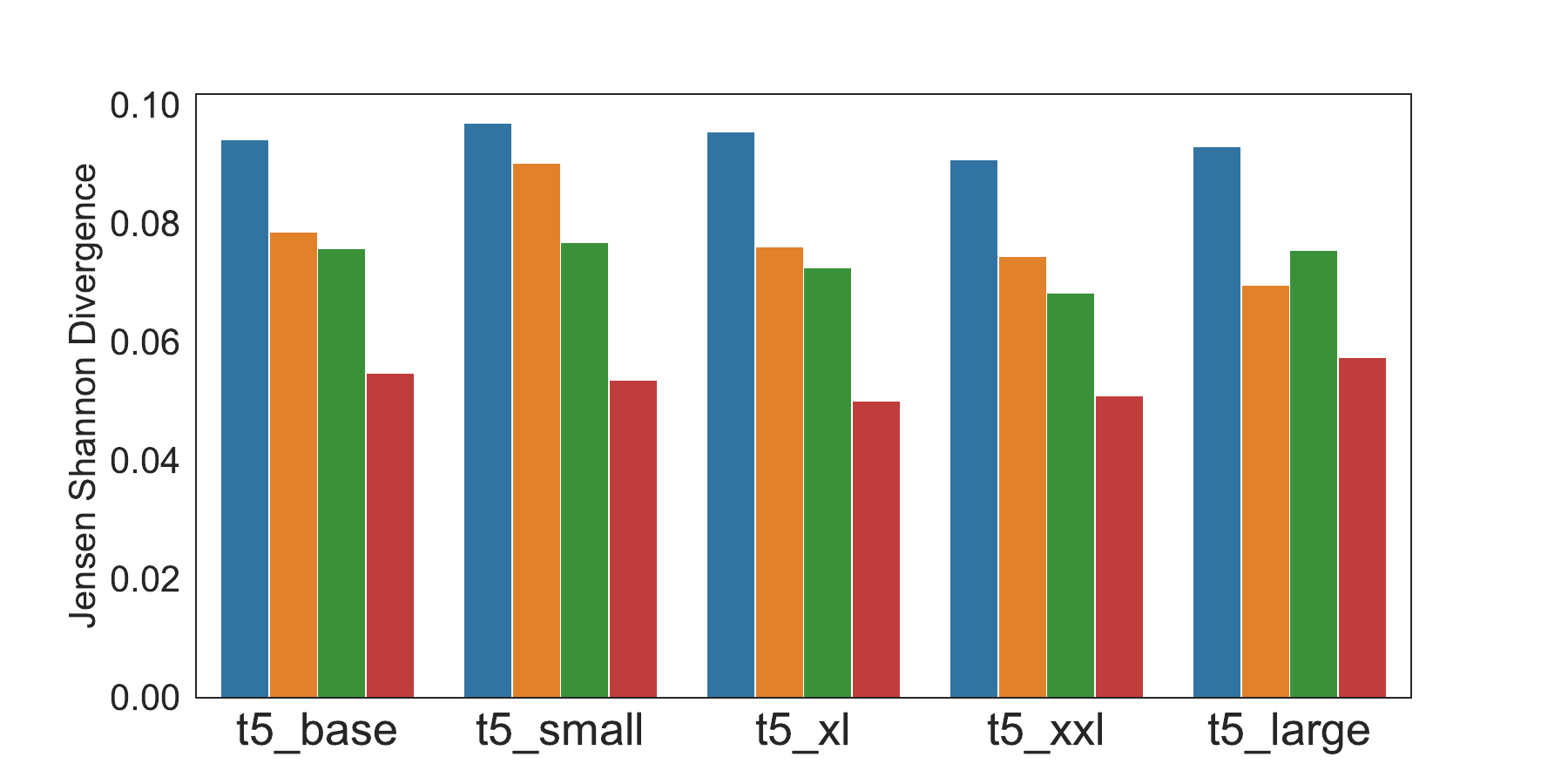}} \\
% \subfloat[caption]{\includegraphics[width = 1.5in]{figures/fingerprint_grid/glm_jsd.pdf}}\\
\subfloat[LLaMA]{\includegraphics[width = 2 in]{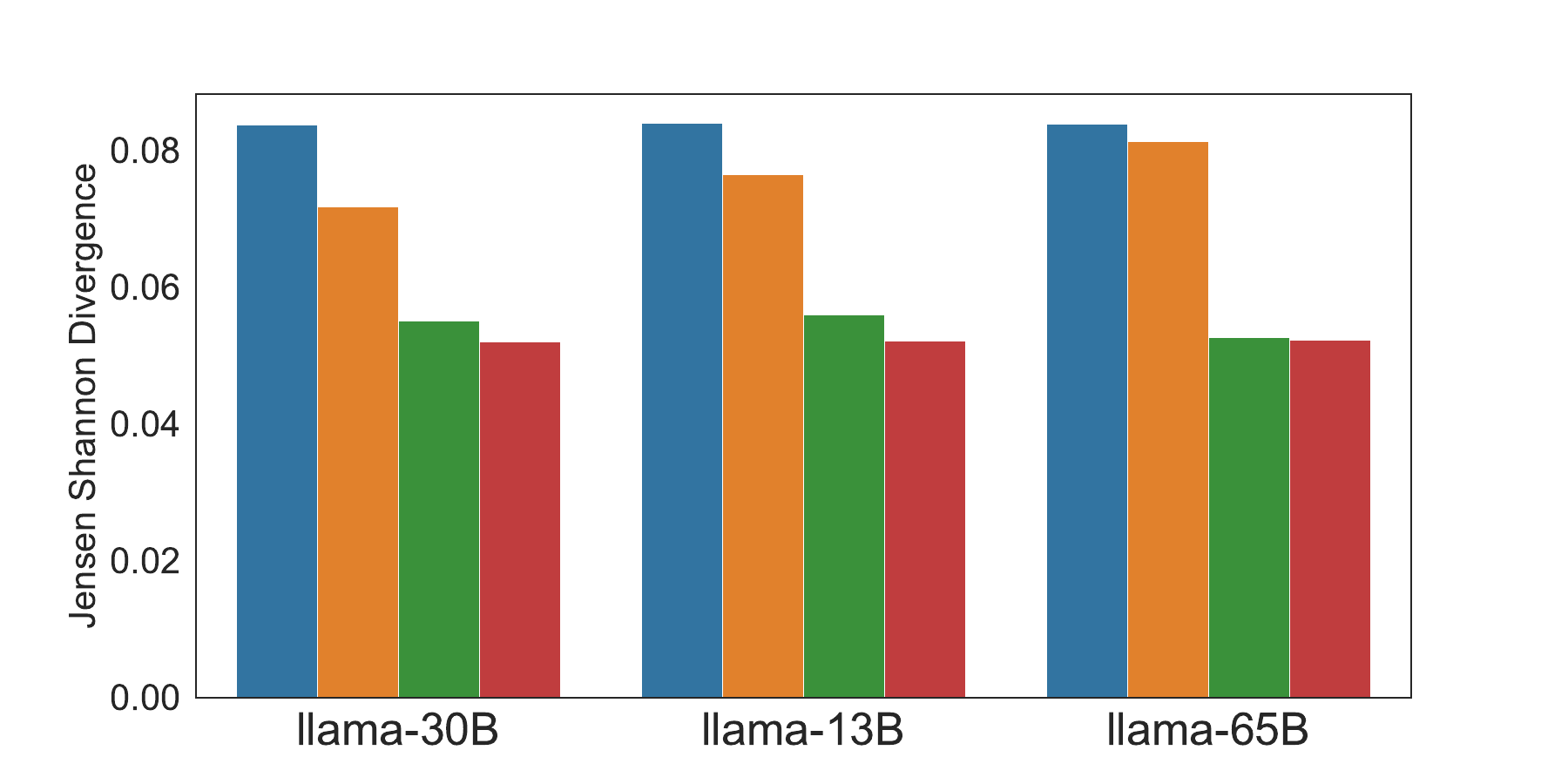}} &
\subfloat[OpenAI]{\includegraphics[width = 2 in]{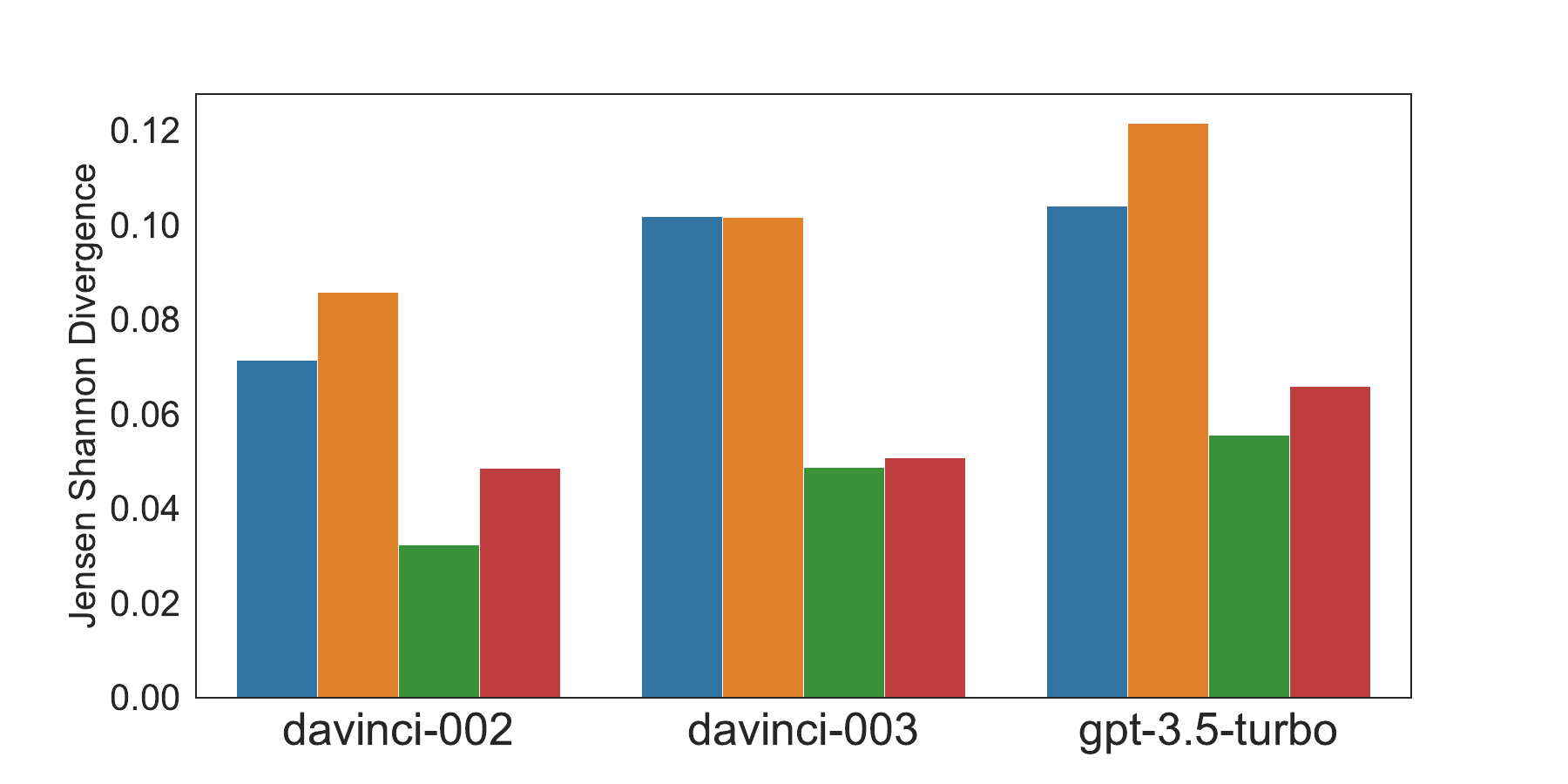}} &
\subfloat[OPT]{\includegraphics[width = 2 in]{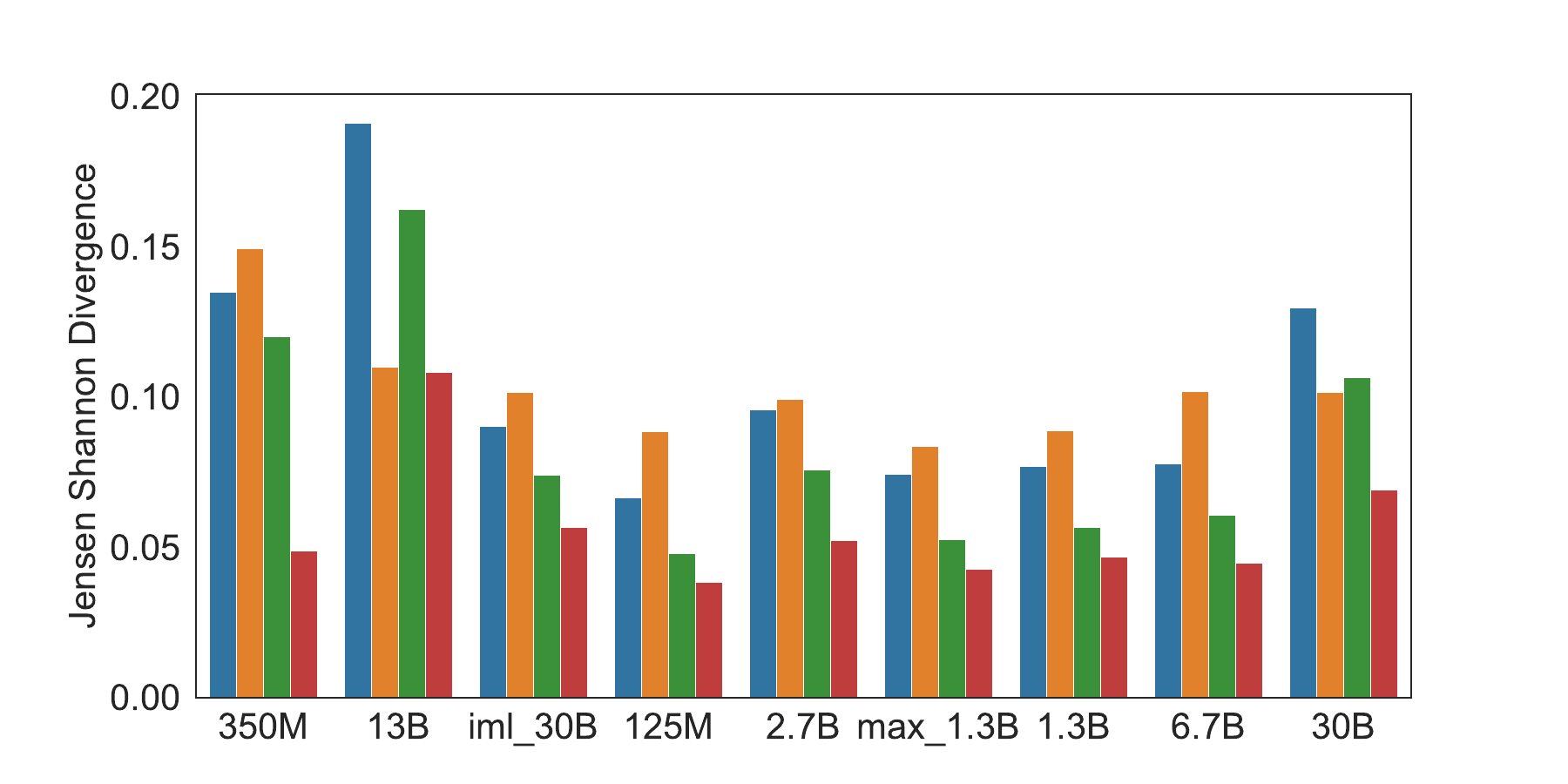}} 
% \subfloat[caption]{\includegraphics[width = 1.5in]{figures/fingerprint_grid/flan_jsd_outfox.pdf}}\\
\end{tabular}
\caption{\textbf{Fingerprint characterization of Deepfake data by model and family.} We report the Jensen-Shannon Divergence of human vs. model for each model in each model family in the Deepfake data across four categories. \textbf{Columns from left to right: constituency type, named entity tag, POS tag, top-$k$ word frequency}. We omit the GLM family in this visualization as there is only one model (130B) available. We do include the GLM results in table format in \autoref{tab:fingerprint_deepfake}. Like in \autoref{fig:radial_fingerprints}, some model families exhibit remarkably consistent fingerprints within families, e.g. LLaMa, Flan, and OpenAI. OPT and EleutherAI in particular have less distinguishable fingerprints within family.}
\end{figure}
\begin{figure}
    \centering
    \includegraphics[width=0.5\linewidth]{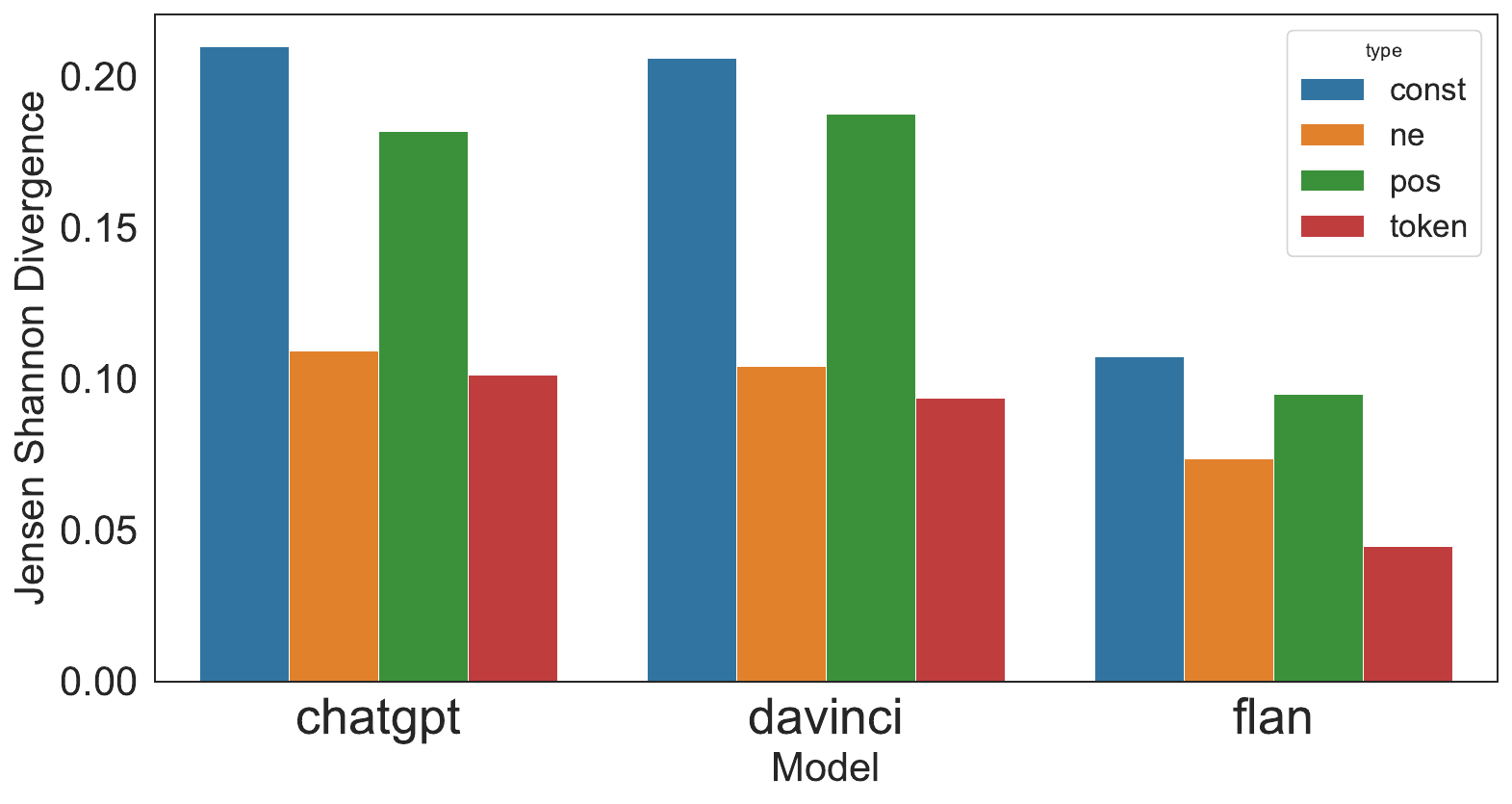}
    \caption{\textbf{Fingerprint characterization of Outfox data by model}. \textbf{Columns from left to right: constituency type, named entity tag, POS tag, top-$k$ word frequency}. We note again that ChatGPT and davinci, being in the same OpenAI model family, have very similar fingerprints, whereas Flan's fingerprint differs substantially. Note that this fingerprint does look different than the Deepfake davinci's fingerprint, showing us that there is some domain dependence to fingerprints, while underscoring the point that regardless of domain, individual models of the same family do produce similar-sounding texts.}
    \label{fig:enter-label}
\end{figure}

\newpage

% \usepackage{booktabs}
\begin{table}
\centering
\footnotesize
\begin{tabular}{@{}ccccc@{}}
\toprule
\textbf{model}       & \textbf{const}    & \textbf{ne}       & \textbf{pos}      & \textbf{top-$k$ token}    \\ \midrule
3.5-turbo   & 0.104062 & 0.121635 & 0.055647 & 0.065913 \\
davinci-002 & 0.071464 & 0.085834 & 0.032347 & 0.048616 \\
davinci-003 & 0.101841 & 0.101650 & 0.048786 & 0.050924 \\ \midrule
t0\_11b     & 0.099334 & 0.116554 & 0.083307 & 0.046207 \\
t0\_3b      & 0.096778 & 0.096023 & 0.081568 & 0.044135 \\
bloom\_7b   & 0.113688 & 0.145603 & 0.096391 & 0.062792 \\ \midrule
llama\_13B  & 0.084033 & 0.076458 & 0.055965 & 0.052204 \\
llama\_30B  & 0.083770 & 0.071776 & 0.055118 & 0.052044 \\
llama\_65B  & 0.083919 & 0.081378 & 0.052662 & 0.052289 \\ \midrule
GLM130B     & 0.092373 & 0.079987 & 0.055011 & 0.058290 \\ \midrule
gpt\_j      & 0.091120 & 0.099940 & 0.079746 & 0.062767 \\
gpt\_neox   & 0.139291 & 0.126545 & 0.098795 & 0.150069 \\ \midrule
iml\_30b    & 0.090410 & 0.101563 & 0.074169 & 0.056793 \\
max\_1.3b   & 0.074518 & 0.083632 & 0.052747 & 0.042955 \\
opt\_1.3b   & 0.076921 & 0.088855 & 0.056709 & 0.046881 \\
opt\_125m   & 0.066651 & 0.088566 & 0.047990 & 0.038409 \\
opt\_13b    & 0.191208 & 0.110141 & 0.162453 & 0.108174 \\
opt\_2.7b   & 0.095768 & 0.099351 & 0.075968 & 0.052447 \\
opt\_30b    & 0.129652 & 0.101543 & 0.106709 & 0.069312 \\
opt\_350m   & 0.134956 & 0.149431 & 0.120094 & 0.049036 \\
opt\_6.7b   & 0.077973 & 0.102048 & 0.060880 & 0.044933 \\ \midrule
t5\_base    & 0.094128 & 0.078521 & 0.075754 & 0.054816 \\
t5\_large   & 0.093037 & 0.069552 & 0.075520 & 0.057391 \\
t5\_small   & 0.096973 & 0.090165 & 0.076765 & 0.053622 \\
t5\_xl      & 0.095449 & 0.076024 & 0.072617 & 0.050033 \\
t5\_xxl     & 0.090850 & 0.074514 & 0.068368 & 0.050967 \\ \bottomrule
\end{tabular}
\caption{\textbf{Jensen Shannon Divergence of human vs. model across fingerprint axes on Deepfake data}. Here we report in full the fingerprint measurements across constituency type (`const'), named entity type (`ne'), part-of-speech (`pos'), and top-$k$ token frequency (`top-$k$ token'). Horizontal lines delineate model families.}
\label{tab:fingerprint_deepfake}
\end{table}
% \usepackage{booktabs}
\begin{table}
\centering
\footnotesize
\begin{tabular}{@{}ccccc@{}}
\toprule
\textbf{model}   & \textbf{const}    & \textbf{ne}       & \textbf{pos}      & \textbf{top-$k$ token}    \\ \midrule
chatgpt & 0.210140 & 0.109395 & 0.181909 & 0.101498 \\
davinci & 0.206403 & 0.104278 & 0.187878 & 0.093806 \\
flan    & 0.107426 & 0.073845 & 0.095178 & 0.044870 \\ \bottomrule
\end{tabular}
\caption{\textbf{Jensen Shannon Divergence of human vs. model across fingerprint axes on Outfox data.} Here we report in full the
fingerprint measurements across constituency type (‘const’), named entity type (‘ne’), part-of-speech (‘pos’), and
top-k token frequency (‘top-k token’).}
\label{tab:fingerprint_outfox}
\end{table}
\section{Additional Experiments}
\label{apx:additional_exp}
% deepfake mixed domains
\begin{table}
\footnotesize
    \centering
    \subfloat[Linguistic feature ablation on Deepfake considering data produced by gpt-j-6b across all domains.]{\begin{tabular}{c|cc}\toprule
     \textbf{Feature Sets} & \textbf{F1} & \textbf{AUROC} \\ \midrule
            character n-grams & \round{2}{0.9583617747440273}  &  \round{2}{0.9917691954022989} \\
            word n-grams &\round{2}{0.7735735735735736} & \round{2}{0.8175473563218392}\\
            POS-tag n-grams & \round{2}{0.8763825634352636} & \round{2}{0.9509126436781609}\\ \midrule
            all & 0.95& 0.95\\ \bottomrule 
    \end{tabular}}
    % \caption{all of deepfake}
    \label{tab:ablations}
    \quad 
    \subfloat[Linguistic feature ablation on Outfox data for content generated by Flan T5 in the one domain present in Outfox (student essay).]{\begin{tabular}{c|cc}\toprule
     \textbf{Feature Sets} & \textbf{F1} & \textbf{AUROC} \\ \midrule
            character n-grams & \round{2}{0.962283384301733}  &  \round{2}{0.9951899999999999} \\
            word n-grams &\round{2}{0.8638011393060591} & \round{2}{0.9418424999999999}\\
            POS-tag n-grams & \round{2}{0.881048387096774} & \round{2}{0.9535039999999999}\\ \midrule
            all & 0.95& 0.95\\ \bottomrule 
    \end{tabular}}
    \caption{\textbf{Linguistic Feature ablation.} For two single model subsets of our data, we compare F1 and AUROC for only characater-, word-, or POS n-gram features to a combination of the three. In both cases, each feature set clearly carries some signal, although word n-grams individually perform the worst. This is likely because word-level features are highly influenced by topic and therefore may not be the most useful feature to help the model generalize across text domain. Interestingly, character n-grams by themselves appear to be stronger than the combined feature set. This warrants more exploration, but we posit this is because the range of character n-grams captures high-frequency subword tokens in a given model's vocabulary.}
\end{table}
% outfox flan
% \begin{table}
% \footnotesize
%     \centering
%     \begin{tabular}{c|cc}\toprule
%      \textbf{Feature Sets} & \textbf{F1} & \textbf{AUC} \\ \midrule
%             character n-grams & \round{2}{0.962283384301733}  &  \round{2}{0.9951899999999999} \\
%             word n-grams &\round{2}{0.8638011393060591} & \round{2}{0.9418424999999999}\\
%             POS-tag n-grams & \round{2}{0.881048387096774} & \round{2}{0.9535039999999999}\\ \midrule
%             all & 0.95& 0.95\\ \bottomrule 
%     \end{tabular}
%     \caption{Outfox Flan}
%     \label{tab:ablations}
% \end{table}

\begin{table}[thbp]
\footnotesize
    \centering
    \label{tab:classifier_comparison}
    \begin{tabular}{lrrrr} \toprule
            & \multicolumn{4}{c}{Average drop in performance} \\ \cmidrule(lr){2-5}
    Experiment & HRec & MRec & F1 & AUC \\ \midrule
    Same Family | Different Domain & \round{2}{-0.02938962921819895}& \round{2}{-0.01125126492075316} & \round{2}{-0.023198} & \round{2}{-0.004074} \\
    Different Family | Same Domain &\round{2}{0.0034795722712225785} & \round{2}{-0.6192498252856342}&\round{2}{-0.210575} & \round{2}{-0.438449}\\
        \bottomrule
    \end{tabular}
    \caption{ \textbf{Models exhibit individual writing styles which are more similar across domains than across model families.} We report the average drop in performance of a GradientBoost from a binary classifier trained on Deepfake data. In 7 independent trials, we train a classifier on a randomly selected model and compare its performance on the in-domain test set to: (1) data from a model in the same family but in a held-out domain, and (2) data from a model in a different family but same domains present in the train set (this is made possible by the fact that Deepfake is multi-parallel). Performance drop is low over data from a model in the same family, and high over data from a model in a different family. The human recall value is small but not 0 as the human data is shuffled and downsampled, so the exact same set of prompts is not seen in every trial.}
    \label{tab:ood_v_oom}
\end{table}
\begin{table}
\footnotesize
\centering
\begin{tabular}{lcc}
\toprule
\textbf{Domain}       & \textbf{In-domain}                           & \textbf{As held-out}\\ \midrule
Finance      & \multicolumn{1}{c|}{\round{3}{0.9794648413192284}} & \round{3}{0.7508305647840531}   \\
Medicine     & \multicolumn{1}{c|}{\round{3}{0.9961832061068702}} & \round{3}{0.9413854351687388}   \\ 
Open QA     & \multicolumn{1}{c|}{\round{3}{0.9959514170040485}} & \round{3}{0.8316831683168318}   \\
Reddit eli5 & \multicolumn{1}{c|}{\round{3}{0.999249361957664}}  & \round{3}{0.828968184215152}    \\
Wiki csai   & \multicolumn{1}{c|}{\round{3}{0.956772334293948}}  & \round{3}{0.7390396659707725}   \\ \midrule
\textbf{Average} & 0.985 & 0.818 \\ \midrule
All          & \round{3}{0.9831614396969058}                      & —              \\ 
\bottomrule
\end{tabular}
\caption{\textbf{Out-of-domain F1 results on HC3 data by domain}. For each domain in the left-hand column, we train two models: (1) one where training data is all domains and the held-out test set is in this domain, and (2) one where training data is all domains \textit{except} this domain, and the held-out test set is in this domain. The final row denotes an experiment in which we train of all domains mixed and test on held-out data which is all in-domain. Some combinations of domains are better for generalizing to unseen domains, which is a noted phenomenon in AID as well \cite{Soto2021LearningUA}; however in general, out-of-domain performance on HC3 is not as strong as it is for Deepfake data (\autoref{fig:deepfake_ood}). This could potentially be improved by increasing the number of samples seen during training (which in each of these experiments was 5000)}
\label{tab:hc3_cross_domain}
\end{table}
\begin{table*}
\footnotesize
\centering
\begin{tabular}{lllllll}
\toprule
Model                        & Accuracy & F1                 & AUROC                & HumanRec           & MachineRec         & AvgRec             \\ \midrule
\multicolumn{1}{l|}{ChatGPT} & \round{3}{0.946}    & \round{3}{0.9491525423728813} & \round{3}{0.9963677675371223} & \round{3}{1.0}                & \round{3}{0.8911290322580645} & \round{3}{0.9455645161290323} \\
\multicolumn{1}{l|}{BloomZ}  & \round{3}{0.992}    & \round{3}{0.9921259842519685} & \round{3}{0.9999519969278033} & \round{3}{1.0}                & \round{3}{0.9838709677419355} & \round{3}{0.9919354838709677} \\
\multicolumn{1}{l|}{Dolly}   & \round{3}{0.714}   & 0\round{3}{.7628524046434495} & \round{3}{0.8449020737327189} & \round{3}{0.9126984126984127} & \round{3}{0.5120967741935484} & \round{3}{0.7123975934459805} \\
\multicolumn{1}{l|}{Cohere}  & \round{3}{0.776}   & \round{3}{0.7227722772277229} & \round{3}{0.9378520225294419} & \round{3}{0.5793650793650794} & \round{3}{0.9758064516129032} & \round{3}{0.7775857654889913} \\
\multicolumn{1}{l|}{Davinci} & \round{3}{0.868}    & \round{3}{0.882142857142857}  & \round{3}{0.9770545314900153} & \round{3}{0.9801587301587301} & \round{3}{0.7540322580645161} & \round{3}{0.8670954941116231} \\
\bottomrule
\end{tabular}
\caption{\textbf{Bert features + GradientBoost classifier on M4 data.} Here we take 5000 training sample, extract BERT pre-trained embeddings for them using Huggingface's feature-extractor pipeline, and train a GradientBoost Classifier. We then test on in-domain data from a held out test set of 500 examples. Dolly and Cohere fool the classifier the most; for Dolly, the machine recall value is the cause of its low performance, perhaps indicating high linguistic variation of the model's generation (i.e. it is easier to pinpoint what a `human' text sounds like than what a `machine' text sounds like). The opposite is true for Cohere. Our model is able to find distinctive markers of machine text but the human data is indistinctive.}
\label{tab:m4_bert}
\end{table*}
\begin{table*}[]
\footnotesize
\centering
\begin{tabular}{lllllll}

\toprule
Model                               & Accuracy           & F1                 & AUROC                & HumanRec            & MachineRec         & AvgRec             \\ \midrule
\multicolumn{1}{l|}{ChatGPT } & \round{3}{0.8515 }            & \round{3}{0.8705130068304026} & \round{3}{0.9928626666666668} & \round{3}{0.7046666666666667}  & \round{3}{0.9983333333333333} & \round{3}{0.8514999999999999} \\
\multicolumn{1}{l|}{BloomZ }  & \round{3}{0.9955 }            & \round{3}{0.9955186721991702} & \round{3}{0.9999122222222222} & \round{3}{0.9913333333333333}  & \round{3}{0.9996666666666667} & \round{3}{0.9955 }            \\
\multicolumn{1}{l|}{Dolly}   & \round{3}{0.7068333333333333} & \round{3}{0.7680949241924852} & \round{3}{0.8749140000000001} & \round{3}{0.44266666666666665} & \round{3}{0.971}              & \round{3}{0.7068333333333333} \\
\multicolumn{1}{l|}{Cohere}  & \round{3}{0.5511666666666667} & \round{3}{0.4207356420735643} & \round{3}{0.6357857777777778} & \round{3}{0.7763333333333333}  & \round{3}{0.326}              & \round{3}{0.5511666666666667} \\
\multicolumn{1}{l|}{Davinci } & \round{3}{0.7673333333333333} & \round{3}{0.8104805864784143} & \round{3}{0.9739818888888888} & \round{3}{0.5396666666666666}  & \round{3}{0.995}              & \round{3}{0.7673333333333333} \\
\bottomrule
\end{tabular}
\caption{\textbf{Linguistic features + GradientBoost classifier on M4 data.} Here we take 5000 training sample, extract linguistic features (char-, word-, POS n-grams) from them, and train a GradientBoost Classifier. We then test on in-domain data from a held out test set of 500 examples. Dolly and Cohere fool the classifier the most; in \autoref{tab:m4_bert}, Dolly exhibits a high machine recall value whereas Cohere exhibits low machine recall. In this case, the opposite is true. This warrants further exploration. Possibly, a combined approach of neural and linguistic features such as in \citet{Petukhova2024PetKazAS} would be more robust here.}
\end{table*}

\begin{table*}
\footnotesize
    \centering
    \begin{tabular}{cccccccc} \toprule
    \small Dataset &  Base Model &  Domain &  F1 & AUROC & HumanRec & MachineRec & AvgRec \\ \midrule
         Domain-Specific & gpt-j-6b & cmv&  0.98&  0.98&0.97&1.00&0.98\\
                                            & & eli5&  0.94&  0.94&0.98&0.90&0.94\\
                                            & & hswag&  0.96&  0.98&1.00&0.97&0.98\\
                                            & & roct& 0.99&  0.99&1.00&0.98& 0.99\\
                                            & & sci\_gen& 0.99 & 0.99&0.99&0.98&0.99\\
                                            & & squad&  0.91&  0.97&1.00&0.95&0.97\\
                                            & & tldr&  0.97&  0.96&0.97&0.95&0.96\\
                                            & & xsum&  0.95&  0.95&0.95&0.95&0.95\\ 
                                            && yelp &0.95& 0.94&0.99&0.89&0.94\\
                                            && wp & 0.98 & 0.98&1.00&0.96&0.98\\ \midrule
                                      % & & \textbf{Average} & \round{2}{0.962}  & \round{2}{0.968} & \round{2}{0.985} & \round{2}{0.953} & \\ \midrule
        Mixed Domains & gpt-j-6b & mixed & 0.95 & 0.95 &0.97&0.93&0.95\\
        & gpt-3.5-turbo & mixed& \round{2}{0.9029850746268657} & \round{2}{0.9655256183745583} & \round{2}{0.9075} & \round{2}{0.909452296819788} & \round{2}{0.908476148409894} \\ 
        & flan-t5-xxl & mixed & \round{2}{0.906693207265489} & \round{2}{0.9455590128755365} & \round{2}{0.911} & \round{2}{0.7886266094420601} & \round{2}{0.8498133047210301}\\ 
        & opt 30b &	mixed & \round{2}{0.9774808215788172} &	\round{2}{0.9931629955947137}&	\round{2}{0.9875	} & \round{2}{0.9273127753303965} &	\round{2}{0.9574063876651983}\\
        &llama 65B & mixed	&\round{2}{0.9172129157562515	} & \round{2}{0.9332129032258064} & \round{2}{0.9445} & \round{2}{	0.7526881720430108	} & \round{2}{0.8485940860215053}\\
        &glm 130B	& mixed & \round{2}{0.9412897016361886	} & \round{2}{0.9503775843307944	} & \round{2}{0.978} & \round{2}{	0.7823721436343852} & \round{2}{0.8801860718171926} \\
        & davinci-003	& mixed & \round{2}{0.8165348303138555} &	\round{2}{0.9215047554347826	} &\round{2}{0.8} & \round{2}{	0.8555253623188406	} & \round{2}{0.8277626811594203} \\\midrule
        
        Mixed Model Set & OpenAI GPT & mixed &\round{2}{0.7015242494226328}&\round{2}{0.7322056234981864}& \round{2}{0.6406276362409313} & \round{2}{0.8237836107554417}&\round{2}{0.7322056234981865} \\
        & Meta Llama* & mixed&\round{2}{0.8056234718826405}&\round{2}{0.8176203830316514} & \round{2}{0.7942151250376619} & \round{2}{0.841025641025641} & \round{2}{0.8176203830316515}\\
        & GLM-130B* & mixed &\round{2}{0.8588098016336055}&\round{2}{0.8600312672780309} & \round{2}{0.8910411622276029} & \round{2}{0.829021372328459} & \round{2}{0.8600312672780309}\\
        & Google FLAN-T5*&mixed &\round{2}{0.83465591257968}&\round{2}{0.8501744651206653}& \round{2}{0.7709685171833693}& \round{2}{0.9293804130579614}&\round{2}{0.8501744651206653}\\
        & Facebook OPT* & mixed &\round{2}{0.8592826714305345}&\round{2}{0.8651724364439806} & \round{2}{0.8719843815367452} & \round{2}{0.8583604913512158} & \round{2}{0.8651724364439806}\\
        & BigScience & mixed &\round{2}{0.7927967985771454}&\round{2}{0.811572401774398} & \round{2}{0.7429166666666667} & \round{2}{0.8802281368821293} & \round{2}{0.8115724017743979} \\
        & EleutherAI & mixed &\round{2}{0.9468451242829828}&\round{2}{0.9495735624929198}&\round{2}{0.9596899224806201} & \round{2}{0.9394572025052192} & \round{2}{0.9495735624929197}\\ \midrule%\cmidrule(lr){3-8}
        % & & \textbf{Average} &  & \\ \midrule
        Ghostbuster & gpt-3.5-turbo & Reuters & \round{2}{0.9842931937172775} & \round{2}{0.9849695633658158} &\round{2}{0.9842931937172775} & \round{2}{0.9856459330143541} & \round{2}{0.9849695633658158} \\
        & & essay & \round{2}{0.9775784753363229} & \round{2}{0.9736595970575597} &\round{2}{0.9864253393665159} & \round{2}{0.9608938547486033} & \round{2}{0.9736595970575597}\\
        & & wp & \round{2}{0.9845360824742267} & \round{2}{0.9851568171009486} & \round{2}{0.9896373056994818} & \round{2}{0.9806763285024155} &\round{2}{0.9851568171009486} \\ \midrule %\cmidrule{3-8}
            % & & \textbf{Average} &  & \\ \cmidrule{3-8}
            & claude & Reuters & \round{2}{0.9669811320754716}& \round{2}{0.9641426282051283}&\round{2}{0.9855769230769231} &\round{2}{ 0.9427083333333334}&\round{2}{0.9641426282051282} \\
            & & essay& \round{2}{0.9781021897810218}& \round{2}{0.9779647435897436}& \round{2}{0.9663461538461539} & \round{2}{0.9895833333333334} &\round{2}{0.9779647435897436} \\
            & & wp & \round{2}{0.9501312335958005} & \round{2}{0.952003704538059} & \round{2}{0.9378238341968912}&\round{2}{0.966183574879227} & \round{2}{0.9520037045380592}\\ \midrule %\cmidrule{3-8}
            % & & \textbf{Average} &  & \\ \midrule
        HC3 & gpt-3.5-turbo & eli5 & \round{2}{0.9992699664184552} & \round{2}{0.9992520602792957} & \round{2}{0.9997078586035641} & \round{2}{0.9988020365378856} & \round{2}{0.9992549475707249}\\
                                & & open\_qa & \round{2}{0.9723991507430998}& \round{2}{0.9952598686533437} & \round{2}{0.9621848739495799} & \round{2}{0.9831932773109243} & \round{2}{0.9726890756302521}\\
                                & &wiki\_csai & \round{2}{0.973134328358209} & \round{2}{0.9946430447113197} & \round{2}{0.9644970414201184} & \round{2}{0.9822485207100592} & \round{2}{0.9733727810650887}\\
                                & & medicine & \round{2}{0.9900199600798404} & \round{2}{0.99976} & \round{2}{0.992} & \round{2}{0.988} & \round{2}{0.99}\\
                                & & finance & \round{2}{0.9724655819774718}&\round{2}{0.9943942948387794} & \round{2}{0.9872935196950444} & \round{2}{0.9567979669631512} & \round{2}{0.9720457433290979}\\ \midrule %\cmidrule{3-8}
                                        % & & \textbf{Average} &  & \\ \midrule
        OUTFOX & gpt-3.5-turbo & essay & \round{2}{0.9460737937559129} & \round{2}{0.9430000000000001} & \round{2}{1.0} & \round{2}{0.886} & \round{2}{0.9430000000000001}\\
                                & text-DaVinci-003& essay & \round{2}{0.9775171065493646}&\round{2}{0.977} & \round{2}{1.0} & \round{2}{0.954}& \round{2}{0.977}\\
                                & flan\_t5\_xxl & essay & \round{2}{0.9539406345957011}&\round{2}{0.955} & \round{2}{0.932} & \round{2}{0.978} & \round{2}{0.9550000000000001}\\ %\midrule %\cmidrule(lr){3-8}
                                        % & & \textbf{Average} & 0.96 & 0.96 & 0.98 & 0.94&0.96\\ 
                                        \bottomrule
    \end{tabular}
    \caption{\textbf{Full In-Domain Results with GradientBoost classifier}. Here we report classifier metrics for four datasets, broken down by model, model family, and domain.}
    \label{tab:all_ngrams}
\end{table*}

\textbf{Fingerprints are ``genetic'': often consistent within a model family.}
A particularly interesting finding is that our classifier is better at detecting out-of-domain texts from the same model than it is detecting in-domain texts from a different model. In other words, Flan T5 ``sounds'' like T5 whether it is generating news stories or fan fiction.  In \autoref{fig:ood_v_oom} we report the results of taking a classifier trained on data from one model and looking at a drop in machine recall when evaluated on either an in-domain test from from a different model or an out-of-domain test from the same model. The classifier is far more effective in the latter case, and the 95\% confidence interval computed with bootstrapping tells us that this is not due to particular oddities of the data splits we used, but that is it a meaningful result. 

We also explicitly test how well a classifier trained on one model generalizes to (1) other models in the same family and (2) other model families. We find that, on average, the drop in machine recall value (out of 1) from in-domain data to other models in the same family is only 0.01, while the drop to other families is 0.62. We report these results in \autoref{tab:ood_v_oom}.

\textbf{AID methods are generally effective for LLM text detection in publicly available datasets.}
\begin{table}[]
\footnotesize
\centering

\subfloat[\textbf{Domain-Specific results:} AUROC for binary classification where ``model'' data is from a single model in a single domain. Where we have data for more than one domain per model, we average them, reporting mean and standard deviation. Outfox data only has one domain (essay), so there is no standard deviation to report.]{\begin{tabular}{lll}
        \toprule
        \textbf{Dataset}                 & \textbf{Model}   & \textbf{AUROC}   \\ \midrule
        Deepfake   & gpt-j-6b         & 96.8 $\pm$ 1.8              \\ \midrule
        HC3        & gpt-3.5-turbo    & 99.6 $\pm$ 0.46              \\ \midrule
        Ghostbuster & gpt-3.5-turbo    & 98 $\pm$ 0.81                \\
                     & claude           & 96.3 $\pm$ 1.2              \\ \midrule
        Outfox      & gpt-3.5-turbo    & 94.0                       \\
                    & text-davinci-003 & 98.0                      \\
                    & flan-t5-xxl      & 96.0                       \\ 
        \bottomrule
        
    \end{tabular}}
    % \label{tab:summary_domain_specific}
\quad
\subfloat[\textbf{Mixed Domain results:} For those datasets for which we have data from a single model in multiple domains, we report AUROC on a held-out test set.]{
\begin{tabular}{llc}
\toprule
\textbf{Dataset}  & \textbf{Model}   & \textbf{AUROC} \\ \midrule
Deepfake & gpt-3.5-turbo & 0.97  \\
         & flan-t5-xxl   & 0.95  \\
         & opt 30B       & 0.99  \\
         & llama 65B     & 0.93  \\
         & glm 130 B     & 0.95   \\
         & davinci-003   & 0.92  \\  \midrule
M4       & gpt-3.5-turbo & 0.99  \\
         & BloomZ        & 1.0   \\
         & Dolly         & 0.88  \\
         & Cohere        & 0.64  \\
         & Davinci       & 0.97  \\ \midrule
HC3      & gpt-3.5-turbo & 1.0  \\
\bottomrule

\end{tabular}}

\caption{Single and mixed domain results with linguistic features and GradientBoost classifier for our datasets of interest.}
% \label{tab:summary_mixed_domain}
\label{tab:single_and_mixed_domain}
\end{table}

\begin{table}[]
\footnotesize
\centering
\begin{tabular}{llc}
\toprule
\textbf{Dataset}  & \textbf{Model}   & \textbf{AUROC} \\ \midrule
Deepfake & gpt-3.5-turbo & 0.97  \\
         & flan-t5-xxl   & 0.95  \\
         & opt 30B       & 0.99  \\
         & llama 65B     & 0.93  \\
         & glm 130 B     & 0.95   \\
         & davinci-003   & 0.92  \\  \midrule
M4       & gpt-3.5-turbo & 0.99  \\
         & BloomZ        & 1.0   \\
         & Dolly         & 0.88  \\
         & Cohere        & 0.64  \\
         & Davinci       & 0.97  \\ \midrule
HC3      & gpt-3.5-turbo & 1.0  \\
\bottomrule

\end{tabular}
\caption{\textbf{Classifier performance on data produced by a single model in a mixture of domains.} We report classifier (linguistic + GradientBoost) performance as AUROC for Deepfake, M4, and HC3 datasets, which all include data from one model across a variety of domains (Outfox only includes the essay domain). The classifier proves broadly robust, except for Cohere data. These results are extracted from the full table in \autoref{tab:all_ngrams}.}
% \label{tab:summary_mixed_domain}
\end{table}

\begin{table}[]
\footnotesize
\centering
\begin{tabular}{lcc}
\toprule
\textbf{Model}                        & \textbf{Ling+GB}            & \textbf{BERT+GB}  \\ \midrule
\multicolumn{1}{l|}{ChatGPT} & \round{3}{0.9928626666666668} & \round{3}{0.9963677675371223}  \\
\multicolumn{1}{l|}{BloomZ}  & \round{3}{0.9928626666666668} & \round{3}{0.9999519969278033} \\
\multicolumn{1}{l|}{Dolly}   & \round{3}{0.8749140000000001} & \round{3}{0.8449020737327189} \\
\multicolumn{1}{l|}{Cohere}  & \round{3}{0.6357857777777778} & \round{3}{0.9378520225294419} \\
\multicolumn{1}{l|}{Davinci} & \round{3}{0.9739818888888888} & \round{3}{0.9770545314900153} \\
\bottomrule
\end{tabular}
\caption{\textbf{Linguistic vs. BERT features}. We report AUROC for binary classification on M4 data, comparing linguistic and BERT features with the same machine learning classifier. For most models, the performance is comparable, with BERT features slightly edging out linguistic features. Cohere is the outlier, proving to be a difficult classification task for the linguistic  set to capture. However, BERT features remain robust to Cohere data.}
\label{tab:summary_m4}
\end{table}

% it does well in the simple case of binary, one model/one domain
In the most simplistic setup (cf. Table \autoref{tab:single_and_mixed_domain}), binary classification of ``human'' or ``machine'' where data is from one model in one text-domain, our GradientBoost classifier with linguistic features achieves upwards of 0.94 AUROC on every test. 

% in the case of one model/mixed domains
We also compare the classifier performance on data produced by a single model, but in multiple domains mixed, seen in \autoref{tab:single_and_mixed_domain}. While performance, in general, is robust, M4's Cohere data is a clear outlier, only achieving 0.64 AUROC, perhaps indicating that it shows a high degree of linguistic diversity across domain-specific generations. However, in a comparison of linguistic vs. neural features on the M4 dataset, we find that the GB classifier can achieve high performance on Cohere (cf. \autoref{tab:summary_m4}).

% in the case of multiclass, one domain
In a multi-class setup (\autoref{tab:multiclass}), where the classifier must distinguish between ``human'', ``model 1'', and ``model 2'', it achieves an average of 0.94 and 0.91 for the Ghostbuster and Outfox datasets, respectively. 

% in the case of mixed model families, it dips
% Finally, we examine the case of LLM content from model families in multiple domains in Deepfake's data. This 

% generalizing to out-of-domain data
All of the above are calculated using unseen data from the same distribution as the training data, but we see strong out-of-domain performance as well in \autoref{fig:deepfake_ood}. Choosing the largest model in each model family available in Deepfake's data, we calculate the F1 score for both an in-domain and out-of-domain test set. Through multiple independent trials and our bootstrapped confidence intervals, we see that F1 on unseen domains has a larger variance, but that overall the difference in F1 score for in-domain and out-of-domain performance is not statistically significant because the confidence intervals overlap. In other words, linguistic features + ML classifiers do generalize well to data in unseen domains.

% \begin{figure}
%     \centering
%     \includegraphics[width=0.8\linewidth]{figures/llama_7b_boxplot.png}
%     \caption{Logits for Llama-7b on data generated from different base models}
%     \label{fig:llama-logits}
% \end{figure}

% commenting out bc I included the dipper and outfox results in the main paper as a table, this figure is not especially providing any new information
% \begin{figure}
%     \centering
%     \includegraphics[width=0.8\linewidth]{figures/dipper_vs_outfox.png}
%     \caption{GradientBoost classifier applied to data that has been adversarially attacked by DIPPER and OUTFOX.}
%     \label{fig:enter-label}
% \end{figure}

\section{Implementation Details}
\label{apx:implementation}
\subsection{GradientBoost}
Parameters: learning rate of 0.2, number of estimators 90, max depth of 8, max features 'sqrt', sumsample ratio 0.8, random state 10, minimum samples leaf 30 and minimum samples to split 400, these hyperparameters were optimized using Sklearn's gridsearch function. Features: char n-grams:(2,4), word n-grams:(3,5), pos n-grams:(3,5). Maximum 2000 features for each feature set.
\section{Dataset Information}
\label{apx:dataset_info}
\begin{table*}[h]
\footnotesize
    \centering
    \begin{tabular}{ccccc} \toprule
    \small Dataset &  Base Model/Family &  Domain &  Human & Machine\\ \midrule
         Domain-Specific & gpt-j-6b & cmv&  509&  636\\
                                            & & eli5&  952&  863\\
                                            & & hswag&  1000&  868\\
                                            & & roct&  999&  833\\
                                            & & sci\_gen&  950 &  529\\
                                            & & squad&  686&  718\\
                                            & & tldr&  772&  588\\
                                            & & xsum&  997&  913\\ 
                                            && yelp & 984& 856\\
                                            && wp & 940 & 784\\\cmidrule(lr){3-5}
                                      & & \textbf{Total} & 8789 & 7588\\ \midrule
        Mixed Model Set & OpenAI GPT & mixed &67k&67k \\
        & Meta Llama & mixed&37k&37k\\
        & GLM-130B & mixed &9k&9k\\
        & Google FLAN-T5&mixed &47k&47k \\
        & Facebook OPT & mixed &80k&80k\\
        & BigScience & mixed &27k&27k \\
        & EleutherAI & mixed &14k&14k\\\cmidrule(lr){3-5}
        & & \textbf{Total} & 282k & 282k\\ \midrule
        Ghostbuster & gpt-3.5-turbo & Reuters & 500 & 500 \\
                & & essay & 1000 & 1000 \\
                & & wp & 500  & 500\\\midrule
                & & \textbf{Total} & 2000 & 2000\\ \midrule
        HC3 & gpt-3.5-turbo & eli5 & 17.1k & 17.1k\\
                                & & open\_qa & 1.19k&1.19k\\
                                & &wiki\_csai & 842&842\\
                                & & medicine & 1.25k&1.25k\\
                                & & finance & 3.93k&3.93k\\\midrule
                                        & & \textbf{Total} & 24.3k & 24.3k\\ \midrule
        OUTFOX & gpt-3.5-turbo & essay & 15k & 15k\\
                                & text-davinci-003& essay & 15k&15k\\
                                & flan\_t5\_xxl & essay & 15k&15k\\\cmidrule(lr){3-5}
                                        & & \textbf{Total} & 46k & 46k\\ 
                                        \bottomrule
    \end{tabular}
    \caption{Dataset statistics (number of documents) for publicly available machine-generated text detection datasets.}
    \label{tab:data_splits}
\end{table*}
\subsection{Outfox}
Outfox is a parallel human-machine dataset built on the Kaggle Feedback Prize dataset \cite{feedback-prize-effectiveness} and contains approximately 15,000 essay problem statements and human-written essays, ranging in provenance from 6th to 12th grade native-speaking students in the United States. For each problem statement, there is also an essay generated by each of three LLMs: ChatGPT (gpt-3.5-turbo-0613), GPT-3.5 (text-davinci-003), and Flan (FLAN-T5-XXL). Each example contain an instruction prompt (``Given the following problem statement, please write an essay in 320 words with a clear opinion.''), a problem statement (``Explain the benefits of participating in extracurricular activities and how they can help students succeed in both school and life. Use personal experiences and examples to support your argument.''), the text of the essay, and a binary label for human or machine authorship.

While we conduct fingerprint analysis on the whole dataset, we use only the human-written subset of the Outfox data as a training corpus for our fine-tuning setup; given an instruction prompt and problem statement, we fine-tune our LLMs of interest to produce text which minimises cross-entropy loss when compared with the original human-written response to the same problem statement. We withhold a test-set of human-written examples from training to be used for evaluation.
\subsection{Ghostbuster}
\citet{Verma2023GhostbusterDT} provide three new datasets for evaluating AI-generated text detection in creative writing, news, and student essays. Using prompts scraped from the subreddit \texttt{r/WritingPrompts}, the Reuters 50-50 authorship identification dataset, and student essays from the online source IvyPanda, they obtained ChatGPT- and Claude-generated responses and made efforts to maintain consistency in length with human-authored content in each domain.
\subsection{HC3}
We also analyze data from \cite{hc3}, which includes questions from publicly available datasets and wiki sources with human- and ChatGPT-generated responses based on instructions and additional context. The resulting corpus comprises 24,322 English and 12,853 Chinese questions, of which we only use the English split.
\subsection{Deepfake}
The Deepfake corpus is a comprehensive dataset designed for benchmarking machine-generated content detection in real-world scenarios \cite{li_deepfake_2023}. It contains approximately 9,000 human examples across 10 text domains, each paired with machine outputs from 27 models (e.g. GPT-3.5-turbo, text-davinci-002) from 7 different model families (e.g. OpenAI), producing several testbeds designed for examining a detector's sensitivity to model provenance and text domain. Each example contains the text, binary label denoting human or machine, and the source information -- which domain, model, and prompting method were used.  

We reserve the whole Deepfake dataset as an evaluation corpus to allow us to examine the robustness of our proposed methods across different models, model families, and domains. For generation, we take the first 30 tokens (as split by whitespace) of an example as the problem statement and prepend it with a simple continuation instruction prompt: ``Read the following instruction and generate appropriate text. \#\#\# Continue the following text:''.

\textbf{Training Data.} We primarily use the Deepfake and Outfox data for training classifiers to analyze different aspects of the LLM fingerprints. They are both conveniently multi-parallel: they contain N model responses for each human text sample in the dataset. This has the benefit of removing some uncertainty from our classifier results. Performance on the human class is often identical across trials, as the human data is often identical. This allows a controlled test of how our classifier deals with the machine text samples. Additionally, the different testbeds provided in Deepfake provide convenient, parallel domain and model (/model family) data splits. Specifically, we use the mixed model sets and model-specific, domain-specific testbeds from Deepfake.

% \subsection{Hyperparameters}
% \label{sec:llm_hyperparams}
% \textbf{Also put relevant hyperparams in a table here}

% \section{N-Gram Analysis}
\label{sec:appendix}
% \input{tables/ablations}
% \begin{lstlisting}
% from sklearn.feature_extraction.text import CountVectorizer
% from nltk.tokenize import word_tokenize
% class PosVectorizer(CountVectorizer):
%     def fit(self, raw_documents, y=None):
%         X_pos = self.transform_to_pos(raw_documents)return super().fit(X_pos, y)

%     def fit_transform(self, raw_documents, y=None):
%         X_pos = self.transform_to_pos(raw_documents)return super().fit_transform(X_pos, y)

%     def transform(self, raw_documents):
%         X_pos = self.transform_to_pos(raw_documents)return super().transform(X_pos)
    
%     def transform_to_pos(self, corpus):
%         def get_pos_tokens(text):
%             tokens = word_tokenize(text)
%             tagged_tokens = pos_tag(tokens)return " ".join([tag for _, tag in tagged_tokens])
    
%         X_pos = [get_pos_tokens(text) for text in corpus]
%         return X_pos}
% \end{lstlisting}

\section{RLHF Negative Result}
\label{apx:rlhf}
\subsection{Related Works}
\textbf{Watermarking.} Unlike the work of \citet{watermark} on watermarking LLMs, the fingerprints we describe are not intentionally added after-the-fact in order to make a model more detectable, but are naturally imprinted on the model during its training/fine-tuning process.
\textbf{Adversarial Attacks.}
While this work is related to adversarial research in that we attempt to create data which will degrade the performance of a classifier, we have a distinctly different goal than that of adversarial attacking; we are not interested in fooling a detector so much as we are exploring what it is that keeps machine-generated data from being truly indistinguishable from that of humans. 
A variety of adversarial attacks have proved incredibly effective against AI-text detectors \cite{Cai2023EvadeCD, outfox}. 
Recognizing the shortcomings of existing detection methods, some have proposed creating more robust classifiers by augmenting training corpora with more challenging examples.
Recently, \citet{outfox} proposed using perturbed (e.g. paraphrased) text to augment training data.
% finally, we use them in a proposed RLHF setup to remove fingerprints. In this case, 
\textbf{Reinforcement Learning.} Having uncovered evidence of unique LLM writing styles, we designed an experimental setup to try to remove the fingerprints with Reinforcement Learning with Human Feedback (RLHF). The results of this experiment were inconclusive, but we report it in summary in \autoref{sec:discussion}. For this setup, we use a mixture of Deepfake and Outfox data for both instruction fine-tuning and RLHF training data. 

\subsection{Generation}
For some experiments, we use LLaMA models and their instruction-tuned counterparts to generate machine responses to human prompts in the Outfox and Deepfake datasets. Outfox's examples are already formatted with an instruction prompt, (e.g. ``Given the following problem statement, please write an essay in 320 words with a clear opinion.'') and explicit problem statement (e.g.``Explain the benefits of participating in extracurricular activities and how they can help students succeed in both school and life. Use personal experiences and examples to support your argument.''), whereas Deepfake's are not. To remedy this, for every example in Deepfake, we take the first 30 tokens (as split by whitespace) of the example text as the problem statement and prepend it with a simple continuation instruction prompt: ``Read the following instruction and generate appropriate text. \#\#\# Continue the following text:''. Further information about generations, including model sampling parameters, may be found in \Cref{sec:appendix}.
We conduct minimal pre- and post-processing on our input as well as the generated outputs, only removing newlines and cleaning up whitespace. 

\subsection{Negative Results}
Here we describe an experimental setup devised to remove a given model's fingerprint with reinforcement learning.

% RLHF intro
RLHF is a popular training methodology that uses reinforcement learning principles to encourage models to produce text with desired attributes. 
Specifically, it optimizes model behavior with human feedback, often facilitated by crowd-sourced annotators or domain experts \cite{rlhf}. Direct Preference Optimization (DPO) \cite{rafailov_direct_2023} is a recent training paradigm for implementing RLHF, which eliminates the need for fine-tuning a reward model and instead directly uses the language model for optimizing the reward function by solving a classification problem on human preference data.

For three recent LLMs (LLaMA-7b, LLaMA-13b, and Falcon-7b), we performed a supervised fine-tuning step to adapt them to the domain of the training data, which is student-generated essays from Outfox. 
Once the SFT models are trained, we use them as a starting point for DPO training.\footnote{We do this because best practices for DPO from \citep{rafailov_direct_2023} suggest that models should first be adapted to the text domain via fine-tuning before being trained with reinforcement learning.}

In a DPO setup, the model is exposed to a triplet of \texttt{prompt}, \texttt{chosen\_response} and \texttt{rejected\_response} where the model learns to optimize the marginal reward of the chosen response over the rejected response. 
% Rather than using the original human and machine pairs in the training data to create the triplets, we use the SFT model to generate new machine responses for each prompt in Outfox. Intuitively, this is making the problem more difficult for the model, which must choose between two fluent, coherent, on-topic responses, and must therefore learn subtler differences between the two texts.
% what we expect to see
Intuitively, if the SFT or DPO process has been successful in removing the LLM fingerprint, we expect to see classifier performance drop -- the machine responses are more difficult to distinguish from human responses because they exhibit more similar distributions of linguistic features than the responses of the base model do. 
To quantify our results, we used our GradientBoost classifier, a finetuned Longformer \cite{longformer} model released by \citet{li_deepfake_2023}, and DetectGPT as a zero-shot detection method \cite{mitchell_detectgpt_2023}.

% why it's negative
After significant tuning, retraining, and optimizing, we did not consistently see an appreciable decrease in classifier performance after the successive stages of fine-tuning.

\subsection{Limitations of RLHF}
We acknowledge that using RLHF to align model outputs with human data necessarily amplifies any subconscious biases which may be present in the human training data. It is well known that certain demographics are over-represented in the creation of user-generated content on the internet, and as our proposed method focuses exclusively on the syntax and lexical choice of LLMs -- not semantic content -- we therefore have not included any filtering for offensive content. 

\end{document}